\RequirePackage{fix-cm}
\RequirePackage{rotating}
\documentclass[smallextended]{svjour3}       
\smartqed  
\usepackage{graphicx}
\usepackage[compress,numbers]{natbib}

\usepackage{comment}

\usepackage{color}
\definecolor{Orange}{rgb}{1,0.5,0}
\definecolor{Red}{rgb}{1,0,0}
\definecolor{Blue}{rgb}{0,0,1}

\usepackage{bm}
\usepackage{amsmath}
\usepackage{amsfonts} 
\usepackage{amssymb} 
\usepackage{stmaryrd} 

\DeclareMathOperator*{\argmin}{arg\,min}
\newcommand{\RNum}[1]{\uppercase\expandafter{\romannumeral #1\relax}} 

\usepackage[linesnumbered,ruled,vlined]{algorithm2e}

\usepackage{booktabs}
\usepackage{longtable}
\usepackage{array}
\usepackage{multirow}

\usepackage[normalem]{ulem}
\useunder{\uline}{\ul}{}

\usepackage{graphicx}
\usepackage{subfigure}

\usepackage{breakcites}
\usepackage[colorlinks=true,linkcolor=blue,citecolor=blue,breaklinks=true]{hyperref} 

\usepackage{rotating} 

\usepackage{caption}
\usepackage{hyperref}

\begin{document}

\title{
Drug-Target Interaction Prediction via an Ensemble of Weighted Nearest Neighbors with Interaction Recovery
} 
\titlerunning{DTI Prediction via an Ensemble of WkNNIR}        

\author{Bin Liu,  Konstantinos Pliakos, Celine Vens, Grigorios Tsoumakas}

\author{Bin Liu \and Konstantinos Pliakos \and \\
        Celine Vens \and Grigorios Tsoumakas 
}


\institute{Bin Liu (Corresponding Author)\at
              School of Informatics, Aristotle University of Thessaloniki, Thessaloniki 54124, Greece \\
              \email{binliu@csd.auth.gr}           
           \and
           Konstantinos Pliakos \at
              KU Leuven, Campus KULAK, Dept. of Public Health and Primary Care, Kortrijk, Belgium \\
              ITEC, imec research group at KU Leuven \\
            \email{konstantinos.pliakos@kuleuven.be} 
          \and
           Celine Vens \at
          KU Leuven, Campus KULAK, Dept. of Public Health and Primary Care, Kortrijk, Belgium \\
          ITEC, imec research group at KU Leuven \\
          \email{celine.vens@kuleuven.be}
          \and
          Grigorios Tsoumakas \at
          School of Informatics, Aristotle University of Thessaloniki, Thessaloniki 54124, Greece \\
          \email{greg@csd.auth.gr}
}

\date{Received: date / Accepted: date}

\maketitle
\begin{abstract}
Predicting drug-target interactions (DTI) via reliable computational methods is an effective and efficient way to mitigate the enormous costs and time of the drug discovery process. 
Structure-based drug similarities and sequence-based target protein similarities are the commonly used information for DTI prediction. 
Among numerous computational methods, neighborhood-based chemogenomic approaches that leverage drug and target similarities to perform predictions directly are simple but promising ones. 
However, existing similarity-based methods need to be re-trained to predict interactions for any new drugs or targets and cannot directly perform predictions for both new drugs, new targets, and new drug-target pairs. Furthermore, a large amount of missing (undetected) interactions in current DTI datasets hinders most DTI prediction methods.
To address these issues, we propose a new method denoted as Weighted k-Nearest Neighbor with Interaction Recovery (WkNNIR). Not only can WkNNIR estimate interactions of any new drugs and/or new targets without any need of re-training, but it can also recover missing interactions (false negatives). In addition, WkNNIR exploits local imbalance to promote the influence of more reliable similarities on the interaction recovery and prediction processes.
We also propose a series of ensemble methods that employ diverse sampling strategies and could be coupled with WkNNIR as well as any other DTI prediction method to improve performance.
Experimental results over five benchmark datasets demonstrate the effectiveness of our approaches in predicting drug-target interactions. Lastly, we confirm the practical prediction ability of proposed methods to discover reliable interactions that were not reported in the original benchmark datasets.



\end{abstract}

\section{Introduction}
Prediction of drug-target interactions (DTIs) is fundamental to the drug discovery process \cite{Ezzat2018ComputationalSurvey, Chen2018MachinePrediction}. However, the identification of interactions between drugs and specific targets via wet-lab (\textit{in vitro}) experiments is extremely costly, time-consuming, and challenging \citep{Dickson2004KeyDevelopment, Paul2010HowChallenge}. Computational (\textit{in silico}) methods can efficiently complement existing \textit{in vitro} activity detection strategies, leading to the identification of interacting drug-target pairs and accelerating the drug discovery process.

In the past, computational approaches identified DTIs mainly based on known ligands for targets \citep{Jacob2008Protein-ligandApproach} or 3D protein structures \citep{Opella2013StructureSpectroscopy}. 
However, these methods suffer from two limitations. Firstly, they would collapse when the required information (ligand-related information or 3D protein structures) is unavailable. Secondly, they are built on only one kind of information. 
Recently, chemogenomic approaches have attracted extensive interest, because they integrate both drug and target information into a unified framework \citep{Ezzat2018ComputationalSurvey} (e.g. chemical structure information related to drugs and genomic information related to target proteins). Such information is often obtained via publicly available databases. 

Popular chemogenomic approaches rely on machine learning algorithms, which have been widely employed for DTI prediction tasks due to their verified effectiveness \citep{Lo2018MachineDiscovery}. There are several machine learning strategies to handle DTI prediction \citep{Ezzat2018ComputationalSurvey}. Here, we focus on neighborhood approaches. These are typically straightforward and effective methods that are based on similarity functions  \citep{Ding2013Similarity-basedReview}. Albeit rather simple, they are very promising. For instance, neighborhood information is often used to regularize matrix factorization tasks, leading to powerful DTI prediction methods \citep{Liu2016NeighborhoodPrediction}. Drug-drug similarities based on chemical structure and target-target similarities based on protein sequence are the most common types of information employed in DTI prediction tasks \citep{Ezzat2018ComputationalSurvey}.

Many of the existing similarity-based methods follow the transductive learning scheme \citep{Liu2016NeighborhoodPrediction,Mei2013Drug-targetNeighbors,vanLaarhoven2013PredictingProfile,Liang2019ImprovedFactorization,Thafar2020DTiGEMS+:Techniques,Mohamed2020DiscoveringEmbeddings}, where test pairs are presented and used during the training process. Every time new pairs arrive, the learning model has to be re-trained in order to perform predictions for them. This is computationally inefficient and substantially affects the scalability of such models, especially in cases where not all test pairs of drugs and targets can be gathered in advance. On the other hand, an inductive model is built upon a training dataset consisting of a drug set, a target set and their interactivity, and can predict any unseen drug-target pairs without re-training.
Therefore, we employ similarity-based approaches following the inductive learning scheme, which is more flexible and effective to perform predictions for newly arrived test pairs. 

Furthermore, many false negative drug-target pairs are typically included in DTI training sets. These drug-target pairs are actually interacting, but the interactions have not yet been reported (detected) due to the complex and costly experimental verification process \citep{Shi2015PredictingClustering,Ezzat2017Drug-targetFactorization,Buza2017ALADIN:Prediction}. Hereafter, these interactions shall be called \emph{missing interactions}. When we treat the unverified DTIs as non-interacting, we inevitably lose valuable information and introduce noise to the data. Therefore, exploiting possible missing interactions is crucial for the DTI prediction approach \citep{Shi2015PredictingClustering,Pliakos2020Drug-targetReconstruction}.


In DTI prediction, there are four main prediction settings. These include predictions for: unseen pairs of training (known) drugs and targets (S1), pairs of test (new) drugs and training targets (S2), pairs of test (new) targets and training drugs (S3), and pairs of test (new) drugs and test (new) targets (S4). Compared to S1, which only considers pairs consisting of known drugs and targets, the other three settings that focus on predicting interactions for new drugs and/or new targets are more difficult, because they relate to the cold-start problem where interacting information of new drugs (targets) is unavailable \citep{Pliakos2020Drug-targetReconstruction}. This paper focuses on these three (S2, S3, S4) more challenging settings. 

S4 is substantially more arduous than S2 and S3, as the bipartite components of test pairs are new in S4.
Many methods, especially most nearest neighbor based ones \citep{Yamanishi2008PredictionSpaces, vanLaarhoven2013PredictingProfile, Ezzat2017Drug-targetFactorization}, either cannot be applied or show a major drop in prediction performance when it comes to S4. An existing neighborhood-based approach designed for S4 specifically \citep{Shi2018InferringModels}, can not perform predictions in S2 and S3. Therefore, there is a lack of neighborhood-based DTI prediction methods that can successfully handle  each and every one of S2, S3, and S4. Such methods are useful when the prediction setting of interest is unknown at training time.

We address DTI prediction with an emphasis on new drugs (S2), new targets (S3), and pairs of new drugs and new targets (S4). 
First, the formulation of the inductive DTI prediction task that aims to perform predictions for any unseen drug-target pair is defined and its differences with the transductive one are clarified. 
Next, we propose a neighborhood-based DTI prediction method called Weighted k-Nearest Neighbor with Interaction Recovery (WkNNIR). The proposed method can deal with all prediction settings, as well as effectively handle missing interactions. 
Specifically, WkNNIR detects neighbors of test drugs, targets, and drug-target pairs to estimate interactions in S2, S3, and S4, respectively. It updates the original interaction matrix based on the neighborhood information to mitigate the impact of missing interactions. 
In addition, WkNNIR exploits the concept of local class imbalance \citep{Liu2019Synthetic} to weigh drug and target similarities, which boosts interaction recovery and prediction.

Furthermore, we propose three ensemble methods to further improve the performance of WkNNIR and other DTI prediction methods. These methods follow a common framework that aggregates multiple DTI prediction models built upon diverse training sets deriving from the original training set by sampling drugs and targets. They employ three different sampling strategies, namely Ensemble with Random Sampling (ERS), Ensemble with Global imbalance based Sampling (EGS), and Ensemble with Local imbalance based Sampling (ELS). A short preliminary account of these three ensemble methods is given in \citep{liu2020local}.

The performance of the proposed methods is evaluated on five benchmark datasets. The obtained results show that WkNNIR outperforms other state-of-the-art DTI prediction methods. We also show that ELS, EGS and ERS are able to promote the performance of six different base models including WkNNIR in all prediction settings, and ELS is the most effective one. Last, we demonstrate cases where our methods succeed in discovering interactions that had not been reported in the original benchmark datasets. The latter highlights the potential of the proposed computational methods in finding new interactions in the real world. 

The rest of this paper is organized as follows. In Section 2, existing DTI prediction approaches are briefly presented. In Section 3, we define the formulation of inductive DTI prediction and compare it with the transductive one. The proposed neighborhood-based and ensemble methods are presented in Sections 4 and 5, respectively. Experimental results and analysis are reported in Section 6. Finally, we conclude our paper in Section 7.

\section{Related Work}

The fundamental assumption of similarity-based methods is that similar drugs tend to interact with similar targets and vice versa \citep{Kurgan2018SurveyInteractions.}. Similarity-based methods can be divided into four types according to the learning method they employ, namely Nearest Neighborhood (NN), Bipartite Local Model (BLM), Matrix Factorization (MF) and Network Diffusion (ND).

Nearest neighborhood based methods predict the interactions based on the information of neighbors.
Nearest Profile \citep{Yamanishi2008PredictionSpaces} infers interactions for a test drug (target) from only its nearest neighbor in the training set. Weighted Profile (WP) \citep{Yamanishi2008PredictionSpaces} integrates all interactions of training drugs (targets) by weighted average to make predictions. Weighted Nearest Neighbor (WNN) \citep{vanLaarhoven2013PredictingProfile} sorts all training drugs (targets) based on their similarities to the test drug (target) in  descending order and assigns the weight of each training drug (target) according to its rank. In WNN, interactions of training drugs (targets) whose weights are larger than a predefined threshold are considered to make predictions. 
The Weighted k-Nearest Known Neighbor (WKNKN) \citep{Ezzat2017Drug-targetFactorization} was initially proposed as a pre-processing step that transforms the binary interaction matrix into an interaction likelihood matrix, while it can estimate interactions for test drug (target) based on $k$ nearest neighbors of training drugs (targets) as well.
The Similarity Rank-based Predictor \citep{Shi2015SRP:Interactions} predicts interactions for test drug (target) based on the likelihood of interaction and non-interaction obtained by similarity ranking.
All the above NN methods are restricted to the S2 and S3 settings. 
In \citep{Shi2018InferringModels}, three neighborhood-based methods, namely individual-to-individual, individual-to-group, and nearest-neighbor-zone, designed specifically for predicting interactions between test drug and test target (S4) are proposed.
Furthermore, the well-known neighborhood-based multi-label learning method MLkNN \citep{zhang2007ml} is employed for drug side effect prediction \citep{Zhang2015PredictingLearning}. In \citep{Shi2015PredictingClustering}, MLkNN with Super-target Clustering (MLkNNSC) takes advantage of super-targets that are constructed by clustering the interaction profiles of targets.

BLM methods build two independent local models for drugs and targets, respectively. They integrate the predictions of the two models to obtain the final scores of test interactions.
The first BLM method is proposed in \citep{Bleakley2009SupervisedModels}, where a support vector machine is employed as the base local classifier. Furthermore, regularized least squares (RLS) formalized as kernel ridge regression (RLS-avg) and RLS using Kronecker product kernel (RLS-kron) are two other representative BLM approaches that process drug and target similarities as kernels \citep{vanLaarhoven2011GaussianInteraction}. One weakness of the above BLM approaches is that they cannot train local models for unseen drugs or targets. To address this issue, BLM-NII \citep{Mei2013Drug-targetNeighbors}, which is based on RLS-avg, introduces a step to infer interactions for test drugs (targets) based on training data. Analogously, GIP-WNN \citep{vanLaarhoven2013PredictingProfile} extends RLS-kron by adding the WNN process to estimate the interactions for test drugs (targets).
In Advanced Local Drug-Target Interaction Prediction (ALADIN) \citep{Buza2017ALADIN:Prediction}, the local model is a k-Nearest Neighbor Regressor with Error Correction (ECkNN) \citep{Buza2015NearestHubs}, which corrects the error caused by bad hubs. Such hubs are often located in the neighborhood of other instances but have different labels from those instances.
Generally, BLM can predict interactions for test drug-target pairs via a two-step learning process \citep{Schrynemackers2015ClassifyingInference, Stock2018ARegression}, i.e. the interactions between test drugs (targets) and training targets (drugs) are estimated first, then two models based on those predictions are built to estimate interactions between test drugs and test targets.    

MF methods deal with DTI prediction by factorizing the interaction matrix into two low-rank matrices which correspond to latent features of drugs and targets, respectively. Kernelized Bayesian Matrix Factorization with Twin Kernels (KBMF2K) \citep{Gonen2012PredictingFactorization} and Probabilistic Matrix Factorization (PMF) \citep{Cobanoglu2013PredictingFactorization} conduct matrix factorization based on probability theory. KBMF2K follows a Bayesian probabilistic formulation and PMF leverages the probabilistic linear model with Gaussian noise. Collaborative Matrix Factorization (CMF) \citep{Zheng2013CollaborativeInteractions} is applied to DTI prediction via adding low-rank decomposition regularization on similarity matrices to ensure that latent features of similar drugs (targets) are similar as well. In \citep{Liang2019ImprovedFactorization}, soft weighting based self-paced learning is integrated into CMF to avoid bad local minima caused by the non-convex objective function and achieve better generalization. Weighted Graph Regularized Matrix Factorization (WGRMF) \citep{Ezzat2017Drug-targetFactorization} performs graph regularization on latent features of drugs and targets to learn a manifold for label propagation. 
Neighborhood Regularized Logistic Matrix Factorization (NRLMF) \citep{Liu2016NeighborhoodPrediction} combines the typical matrix factorization with neighborhood regularization in a unified framework to model the probability of DTIs within the logistic function. 
Dual-Network Integrated Logistic Matrix Factorization (DNILMF) \citep{Hao2017PredictingFactorization}, is an extension of NRLMF that utilizes diffused drug and target kernels instead of similarity matrices for the logistic function. 
MF approaches usually comply with the transductive learning scheme.

Network-based inference (NBI) \citep{Cheng2012PredictionInference} applies network diffusion on the DTI bipartite network leveraging graph-based techniques to predict new DTIs. This approach uses only the interactions between drugs and targets. Domain Tuned-Hybrid (DT-Hybrid) \citep{Alaimo2013Drug-targetInference} and Heterogeneous Graph Based Inference (HGBI) \citep{Wang2013DrugInference} extends NBI via incorporating drug‐target interactions, drug similarities and target similarities to the diffusion of the heterogeneous network. Apart from network inference, random walk \citep{Chen2012Drug-targetNetwork}, probabilistic soft logic \citep{Fakhraei2014Network-basedLogic}, and finding simple path \citep{Ba-Alawi2016DASPfind:Interactions} are three other approaches that can be applied to the heterogeneous DTI network to predict interactions. All of the ND methods are transductive, as the network diffusion step should be recomputed if new drugs or new targets are added in the heterogeneous network.

Apart from using similarities, there are methods that treat interaction prediction as a classification task, building binary or multi-label classifiers over input feature sets. In \citep{Pliakos2019NetworkTrees}, traditional ensemble tree strategies, such as random forests (RF) \citep{Breiman2001RandomForests} and extremely randomized trees (ERT) \citep{Geurts2006ExtremelyTrees}, are extended to the bi-clustering tree setting \citep{Pliakos2018GlobalPrediction}. Another example is AGHEL \citep{Zheng2018PredictingLearning}, which is a heterogeneous ensemble approach integrating two different classifiers, namely RF and XGBoost \citep{Chen2016XGBoost:System}. Moreover, three tree-based multi-label classifiers which incorporate various label partition strategies to effectively capture the correlations among drugs and targets are proposed for DTI prediction in \citep{Pliakos2019PredictingPartitioning}.
Furthermore, delivering low dimensional drug and target embeddings from a DTI network using graph embedding \citep{Luo2017AInformation,Mohamed2020DiscoveringEmbeddings} or path category based extraction techniques \citep{Olayan2018DDR:Approaches,Chu2019DTI-CDF:Features,Thafar2020DTiGEMS+:Techniques} has been shown very effective. However, these methods are utilized in a transductive setting.

Finally, we briefly present approaches dealing with the issue of missing interactions. The DTI prediction task with missing interactions can be treated as a Positive-Unlabeled (PU) learning problem \citep{Bekker2020LearningSurvey}, where the training data consists of positive and unlabeled instances and only a part of positive instances are labeled. 
Several methods mentioned above discover missing interactions by using matrix completion techniques \citep{Zheng2013CollaborativeInteractions, Liu2016NeighborhoodPrediction}. Others construct super-targets containing more interacting information \citep{Shi2015PredictingClustering}, correct possible missing interactions \citep{Buza2017ALADIN:Prediction}, and recover interactions as a pre-processing step \citep{Ezzat2017Drug-targetFactorization}. 
Similar to the idea of interaction recovery, Bi-Clustering Trees with output space Reconstruction (BICTR) \citep{Pliakos2020Drug-targetReconstruction} utilizes NRLMF to restore the training interactions on which an ensemble of Bi-clustering trees \citep{Pliakos2019NetworkTrees} is built.
In \citep{Shi2016PredictingInteractions,Peng2017ScreeningLearning}, traditional PU learning algorithms, such as Spy and Rocchio, are employed to extract reliable non-interacting drug-target pairs. 
Based on a more informative PU learning assumption that interactions are not missing at random, a probabilistic model without bias to labeled data is proposed in \citep{Lin2019LearningLabels}.

\section{Inductive DTI Prediction}
In this section, the formulation of similarity based DTI prediction in an inductive learning scheme is presented. Then, the comparison between the inductive and transductive settings for the DTI prediction task is given.

\subsection{Formulation}

In this part, we define the inductive DTI prediction problem, as well as the training and estimation procedures of a DTI prediction model in inductive learning, where the drug and target similarities are utilized as input information. 

Let $D=\{d_i\}_{i=1}^n$ be the training drug set containing $n$ drugs, where each drug is a compound described by its chemical structure. Let $T=\{t_i\}_{i=1}^m$ be the training target set consisting of $m$ targets, where each target is a protein represented by its amino acid sequence. There is a set of known interactions between drugs in $D$ and targets in $T$.
Fig.\ref{fig:inductive_train} illustrates the process of training an inductive model.
Initially, to cater for a similarity-based DTI prediction model, the drug similarity matrix $\bm{S}^{d} \in \mathbb{R}^{n \times n}$ and the target similarity matrix $\bm{S}^{t} \in \mathbb{R}^{m \times m}$ are computed, where $S^d_{ij}$ is the similarity between $d_i$ and $d_j$, and $S^t_{ij}$ is the similarity between $t_i$ and $t_j$. In this paper, drug similarities are computed by SIMCOMP algorithm \citep{Hattori2003DevelopmentPathways}, which assesses the common chemical structure of two drugs, and target similarities are calculated by using the normalized Smith-Waterman (SW) score \citep{Smith1981IdentificationSubsequences}, which evaluates the shared amino acid sub-sequence of two targets. In addition, DTIs are represented with an interaction matrix $\bm{Y} \in \{0,1\}^{n \times m}$, where $Y_{ij} = 1$ if $d_i$ and $t_j$ are known to interact with each other and $Y_{ij} = 0$ indicates that $d_i$ and $t_j$ either actually interact with each other but their interaction is undetected, or $d_i$ and $t_j$ do not interact. 
An inductive DTI prediction model is built based on a training set consisting of $D$, $T$, $\bm{S}^d$, $\bm{S}^t$ and $\bm{Y}$.



In the prediction phase, as discussed in the introduction, we distinguish three settings of DTI prediction, according to whether the drug and target involved in the test pair are included in the training set or not. In particular:
\begin{itemize}
    \item S2: predict the interactions between test drugs $\bar{D}$ and training targets $T$.
    \item S3: predict the interactions between training drugs $D$ and test targets $\bar{T}$.
    \item S4: predict the interactions between test drugs $\bar{D}$  and test targets $\bar{T}$.
\end{itemize}
where $\bar{D}=\{ d_u \}_{u=1}^{\bar{n}}$ is a set of test drugs disjoint from the training drug set (i.e. $\bar{D} \cap D = \emptyset$), and $\bar{T}=\{ t_v \}_{v=1}^{\bar{m}}$ is a set of test targets disjoint from $T$.

The prediction procedure is shown in Fig. \ref{fig:inductive_test}. In all prediction settings, the similarities between test drugs (targets) and training drugs (targets), required by the similarity-based model, are firstly computed. Next, the learned DTI prediction model receives these similarities to perform predictions for the corresponding test drug-target pairs.  
In S2, given a set of test drugs $\bar{D}$, the similarities between $\bar{D}$ and $D$ ($\bar{\bm{S}}^d \in \mathbb{R}^{\bar{n} \times n}$) computed by the SIMCOMP
algorithm are input to the model, and the predictions are a real-valued matrix $\hat{\bm{Y}} \in \mathbb{R}^{\bar{n} \times m}$ indicating the confidence of the affinities between test drugs and training targets.
Similarly in S3, the normalized SW score is employed to calculate the  similarities between $\bar{T}$ and $T$ ($\bar{\bm{S}}^t \in \mathbb{R}^{\bar{m} \times m}$ ), upon which the model outputs a prediction matrix $\hat{\bm{Y}} \in \mathbb{R}^{n \times \bar{m}}$.
In S4, both $\bar{\bm{S}}^d$ and $\bar{\bm{S}}^t$ are computed, and the prediction matrix is
$\hat{\bm{Y}} \in \mathbb{R}^{\bar{n} \times \bar{m}}$.

\subsection{Comparison between Inductive and Transductive Settings} 

\begin{figure*}[htp]
\centering
\subfigure[Training phase of an inductive model]{\includegraphics[width=0.6\textwidth]{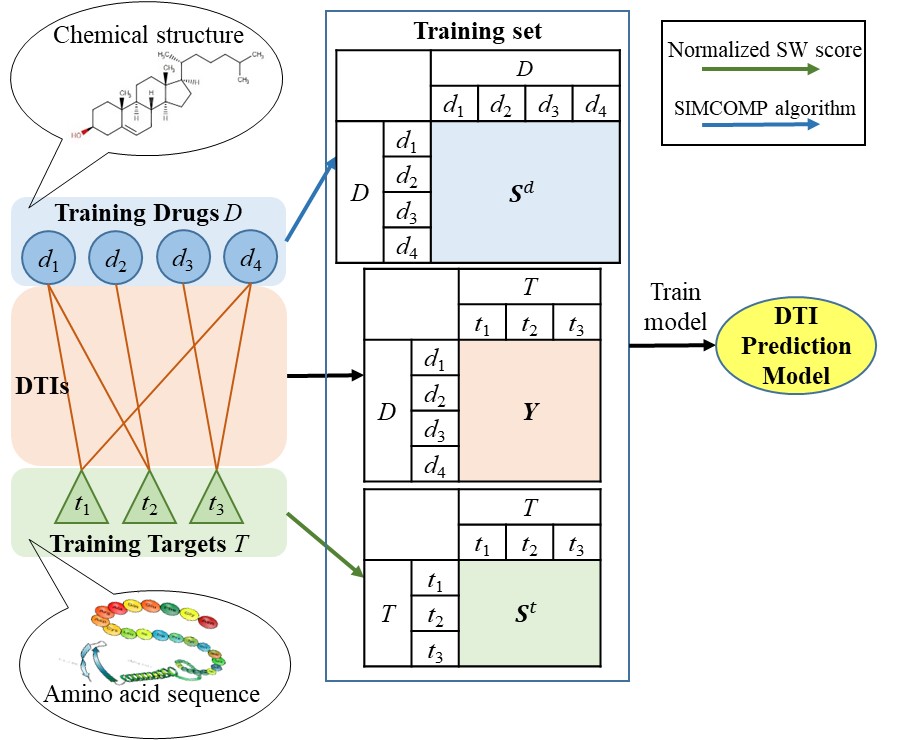}\label{fig:inductive_train}} \\
\subfigure[Prediction phase of an inductive model]{\includegraphics[width=0.7\textwidth]{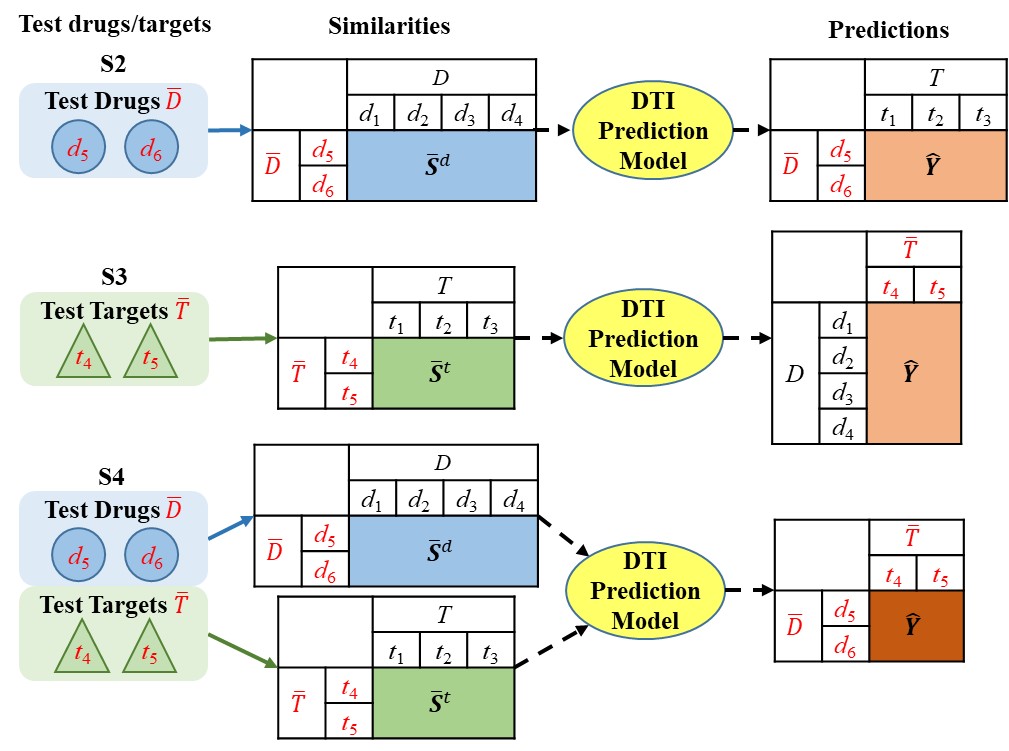}\label{fig:inductive_test}} \\
\subfigure[The training and prediction procedure of a transductive model dealing with S4]{\includegraphics[width=0.9\textwidth]{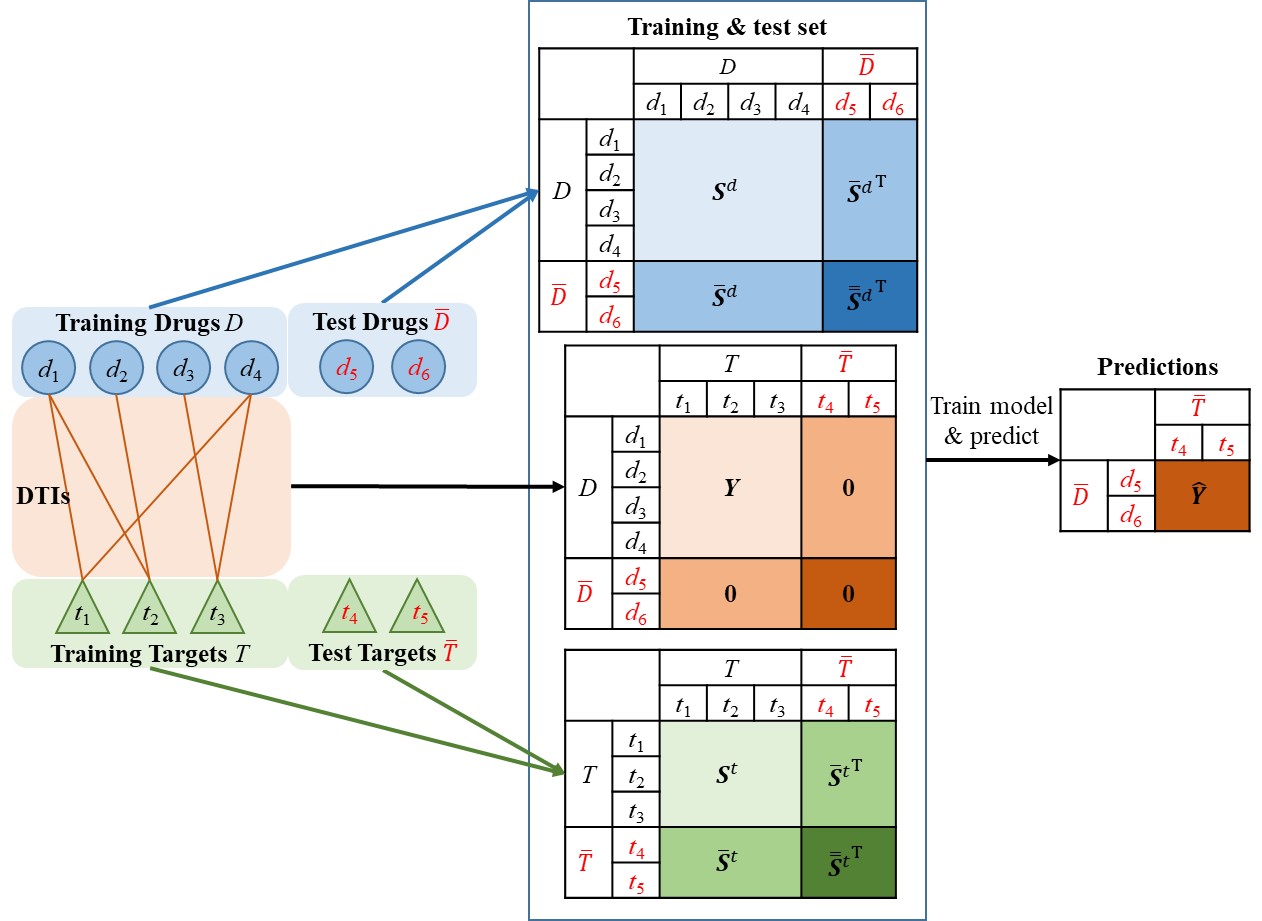}\label{fig:transductive_S4}} 
\caption{Comparison between inductive and transductive settings} 
\label{fig:transductive_inductive}
\end{figure*}

In the transductive learning scheme, the model is learned for the purpose of predicting specific test pairs. Fig. \ref{fig:transductive_S4} describes the training and prediction processes of a model in the transductive setting dealing with S4. Both training and test drugs (targets) are available in the learning phase of a transductive model. Given that $\tilde{D} = D\cup\bar{D}$ and $\tilde{T} = T\cup\bar{T}$, one would first compute the extended drug and target similarity matrices for $\tilde{D}$ and $\tilde{T}$, respectively:
 \begin{equation}
    \tilde{\bm{S}}^d = 
    \begin{bmatrix}
    \bm{S}^d & {\bar{\bm{S}^d}}^\top \\
    \bar{\bm{S}}^d & \bar{\bar{\bm{S}}}^d
    \end{bmatrix}
    \quad
    \tilde{\bm{S}}^t = 
    \begin{bmatrix}
    \bm{S}^t & {\bar{\bm{S}^t}}^\top \\
    \bar{\bm{S}}^t & \bar{\bar{\bm{S}}}^t
    \end{bmatrix}
 \end{equation}
where $\bar{\bar{\bm{S}}}^d \in \mathbb{R}^{\bar{n} \times \bar{n}}$ ($\bar{\bar{\bm{S}}}^t \in \mathbb{R}^{\bar{m} \times \bar{m}}$) stores similarities between test drugs (targets).
In addition, the interaction matrix is extended to $\tilde{{\bm{Y}}} \in\{0,1\}^{(n+\bar{n}) \times (m+\bar{m})}$, where the rows and columns of $\tilde{{\bm{Y}}}$ corresponding to test drugs and targets contain ``0"s.
The transductive model is trained upon the input consisting of $\tilde{D}$, $\tilde{T}$, $\tilde{\bm{S}}^d$, $\tilde{\bm{S}}^t$ and $\tilde{{\bm{Y}}}$, and could perform predictions for the specific test pairs included in the input once it has been built.
The processes of the transductive model handling S2 and S3 are similar with S4, except for that no test target and drug are used in S2 and S3, respectively (e.g in S3, $\bar{D}=\emptyset$, the drug similarity matrix is $\bm{S}^t$ and the interaction matrix is $[\bm{Y}, \bm{0}]$).

The difference between the inductive and transductive settings relies on their input information and the predicting ability of the model.
On one hand, an inductive model is built in the training phase, using similarity and interaction matrices referring to training drugs and targets. After the completion of this process, it can perform predictions for any setting (S2, S3, S4) and any unseen pairs. On the other hand, a transductive model is built upon both the information of the training set and the similarities of the test drugs and/or targets. Hence, it can provide predictions only for these specific test pairs and cannot generalize to unseen test data. If a transductive model needs to predict the interaction of another unseen drug-target pair different to the one included in the training process, it should be re-trained, incorporating the information corresponding to the unseen test pair. This is extremely demanding, especially when it comes to large scale data. Therefore, in this paper, we focus on the inductive setting. 

\section{WKNNIR}

In this section, we first propose the Weighted k-Nearest Neighbor (WkNN) as a comprehensive nearest neighbor based approach that could handle all prediction settings. Next, the local imbalance is introduced to measure the reliability of similarity matrices. Lastly, WkNNIR, which incorporates interaction recovery and local imbalance driven similarity weighing into WkNN, is presented.

\subsection{WkNN}
Neighborhood-based methods predict the interactions of test drugs (S2) or test targets (S3) by aggregating the interaction profiles of their neighbors \citep{Yamanishi2008PredictionSpaces, vanLaarhoven2013PredictingProfile, Ezzat2017Drug-targetFactorization}. 
The major limitation of these methods is that they cannot directly predict interactions between test drugs and test targets (S4), as interactions between test drugs (targets) and their neighbors are unavailable in the training set.   
To overcome this drawback of existing methods, we propose WkNN. 

WkNN employs the prediction function of WKNKN \citep{Ezzat2017Drug-targetFactorization} to deal with S2 and S3 because of its simplicity and efficacy. 
WkNN is a lazy learning method that does not have any specific training phase. Given a test drug-target pair $(d_u,t_v)$ belonging to the prediction setting S2 or S3, WkNN predicts their interaction profiles based on the interactions of either $k$ nearest training drugs of $d_u$ or $k$ nearest training targets of $t_v$ as follows:
\begin{equation}
\hat{Y}_{uv} =\left\{
\begin{aligned}
 & \frac{1}{z_{d_u}}\sum_{d_i \in {\cal N}^k_{d_u}} \eta^{i'-1}\bar{S}^d_{ui}Y_{iv}, \text{ if } d_u \notin D \text{ and } t_v \in T \\
 & \frac{1}{z_{t_v}}\sum_{t_j \in {\cal N}^k_{t_v}}\eta^{j'-1}\bar{S}^t_{vj}Y_{uj}, \text{ if } d_u \in D \text{ and } t_v \notin T
\end{aligned}
\right. 
\label{eq:wknn23}
\end{equation}
In Eq.\eqref{eq:wknn23}, ${\cal N}^k_{d_u}$ (${\cal N}^k_{t_v}$) corresponds to $k$ nearest neighbors of $d_u$ ($t_v$) which are retrieved by choosing $k$ training drugs (targets) having $k$ largest values in $u$-th row of $\bar{\bm{S}}^d$ ($v$-th row of $\bar{\bm{S}}^t$) and sorting the selected $k$ drugs (targets) in the descending order according to their similarities to $d_u$ ($t_v$). 
Moreover, $d_i$ ($t_j$) is the $i'$-th ($j'$-th) nearest neighbor of $d_u$ ($t_v$), i.e. $i'$ ($j'$) is the index of $d_i$ ($t_j$) in ${\cal N}^k_{d_u}$ (${\cal N}^k_{t_v}$), $\eta \in [0,1]$ is the decay coefficient shrinking the weight of further neighbors, and $z_{d_u}=\sum_{d_i \in {\cal N}^k_{d_u}}\bar{S}^d_{ui}$ and $z_{t_v}=\sum_{t_j \in {\cal N}^k_{t_v}}S^t_{vj}$ are normalization terms. 


To make predictions for pairs of test drugs and test targets (S4), WkNN follows the tetrad rule: when a drug interacts with a target, another similar drug probably interacts with another similar target \citep{Fakhraei2014Network-basedLogic}. 
For the sake of adapting to the tetrad rule, WkNN considers similar drug-target pairs instead of similar drugs or targets. We define the neighbors of a test drug-target pair $(d_u,t_v)$ as ${\cal N}^k_{d_ut_v} = \{(d_i,t_j)| d_i \in {\cal N}^k_{d_u},t_j \in {\cal N}^k_{t_v} \}$. Although the direct way to find neighbors of a drug-target pair is to search all $nm$ drug-target combinations, our definition shrinks the search space to $n+m$ leading to increased efficiency. 
As in \citep{vanLaarhoven2011GaussianInteraction, vanLaarhoven2013PredictingProfile}, we define the similarity between two drug-target pairs as the product of the similarity of two drugs and the similarity of two targets, i.e. the similarity of $(d_u,t_v)$ and $(d_i,t_j)$ is $\bar{S}^d_{ui}\bar{S}^t_{vj}$.
Thus, the prediction function of WkNN for S4 is:
\begin{equation}
\hat{Y}_{uv} =\frac{1}{z_{uv}}\sum_{(d_i,t_j) \in {\cal N}^k_{d_ut_v}}\eta^{i'+j'-2}\bar{S}^d_{ui}\bar{S}^t_{vj}Y_{ij}
\label{eq:wknn4}
\end{equation}
where the normalization term $z_{uv}$ is equal to $\sum_{(d_i,t_j) \in {\cal N}^k_{d_ut_v}}\bar{S}^d_{ui}\bar{S}^t_{vj}$.
The weight in Eq.\eqref{eq:wknn4} consists of two parts: the first one corresponds to the decay coefficient, where the exponent is determined by the index of $d_i$ in ${\cal N}^k_{d_u}$ ($i'$) and the index of $t_j$ in ${\cal N}^k_{t_v}$ ($j'$) simultaneously, and the second one is the similarity between the test pair and its neighbor.

\subsection{Local Imbalance}
The concept of local imbalance concerns the label distribution of an instance within the local region, playing a key role in determining the difficulty of a dataset to be learned \citep{Liu2019Synthetic}. 
Concerning DTI data that contain two kinds of similarities, the local imbalance can be assessed in both drug space and target space.

Firstly, we define the drug-based local imbalance. 
The local imbalance of a drug $d_i$ for target $t_j$ is measured as the proportion of ${\cal N}^k_{d_i}$ having the opposite interactivity to $t_j$ as $d_i$:
\begin{equation}
C^d_{ij} = \frac{1}{k}\sum_{d_h \in {{\cal N}^k_{d_i}}} \llbracket Y_{hj} \neq Y_{ij} \rrbracket
\label{eq:Cd_ij}
\end{equation}
where $C^d_{ij} \in [0,1]$ and $\llbracket x \rrbracket$ is the indicator function that returns 1 if $x$ is true and 0 otherwise. The larger the value of $C^d_{ij}$, the fewer the drugs in the local region of $d_i$ having the same interactivity to $t_j$, and the higher the local imbalance of $d_i$ for $t_j$ based on drug similarities. 
In the ideal case, similar drugs share the same interaction profiles (i.e. interact with the same targets), rendering DTI prediction a very simple task. However, there are several cases, where similar drugs interact with different targets, which makes DTI prediction more challenging. 
Therefore, we employ local imbalance $C^d_{ij}$ to assess the reliability of similarities between $d_i$ and other training drugs.
Specifically, lower $C^d_{ij}$ indicates more reliable similarity information.

Likewise, we calculate the local imbalance of $t_j$ for $d_i$  based on the targets as:
\begin{equation}
C^t_{ij} = \frac{1}{k}\sum_{t_h \in {{\cal N}^k_{t_j}}} \llbracket Y_{ih} \neq Y_{ij} \rrbracket
\end{equation}


\subsection{WkNNIR}

The existence of missing interactions (not yet reported) in the training set can lead to biased DTI prediction models and an inevitable accuracy drop.  
To address this issue, we propose WkNNIR, which couples WkNN with interaction recovery to perform predictions upon the completed interaction matrix. 
Moreover, in S4, where both drug and target similarities are used to perform predictions, WkNNIR has the advantage that the importance (weight) of drug and target similarities is differentiated depending on their local imbalance.
 
Firstly, WkNNIR computes recovered interactions, which replace the original interactions in the prediction phase. Based on the assumption that similar drugs interact with similar targets and vice versa, missing interactions can be completed by considering the interactions for neighbor drugs or targets. There are two ways to recover the interaction matrix: one on the drug side and another on the target side. The drug side recovery is conducted row-wise, where each row of $\bm{Y}$ is recovered by the weighted average of the neighbor rows identified by drug similarities. The drug-based recovery interaction matrix $\bm{Y}^d$ is:
\begin{equation}
\bm{Y}^d_{i\cdot} = \frac{1}{z_{d_i}}\sum_{d_h \in {\cal N}^k_{d_i}}\eta^{h'-1} S^d_{ih} \bm{Y}_{i\cdot}, \  i = 1,\cdots,n
\label{eq:Yd}
\end{equation}
where $h'$ is the index of $d_h$ in ${\cal N}^k_{d_i}$, and $z_{d_i}=\sum_{d_h \in {\cal N}^k_{d_i}}S^d_{ih}$.
The target-based recovery interaction matrix $\bm{Y}^t$ is obtained by reconstructing $\bm{Y}$ column-wise:
\begin{equation}
\bm{Y}^t_{\cdot j} = \frac{1}{z_{t_j}}\sum_{t_l \in {\cal N}^k_{t_j}}\eta^{l'-1} S^t_{lj}\bm{Y}_{\cdot j}, \  j = 1,\cdots,m
\label{eq:Yt}
\end{equation}
where $l'$ is the index of $t_l$ in ${\cal N}^k_{t_j}$ and $z_{t_j}=\sum_{t_l \in {\cal N}^k_{t_j}}S^t_{lj}$.

However, $\bm{Y}^d$ ($\bm{Y}^t$) exploits only one kind of similarity and neglects the other one. To address this issue, we combine these two recovered interaction matrices into a complementary one that incorporates the recovery information from both drug and target views.
Besides, interactions restored via more reliable similarity measures tend to be more credible. Therefore, instead of treating $\bm{Y}^d$ and $\bm{Y}^t$ equally, we distinguish the effectiveness of different recovered interaction matrices according to the local imbalance of the similarity used in the recovery process. 

The interacting pair $(d_i,t_j)$ with lower drug-based local imbalance indicates that $d_i$ is close to other drugs interacting with $t_j$ in the drug space, and therefore the recovered interactions inferred from the pair are more reliable in the drug view.
Hence, we define the weight of recovered interaction $Y^d_{ij}$ according to the average local imbalance of interacting pairs that are used to estimate $Y^d_{ij}$:
\begin{equation}
    W^d_{ij} = \exp(-\frac{\sum_{d_h \in {\cal N}^k_{d_i}}C^d_{hj}Y_{hj}}{\sum_{d_h \in {\cal N}^k_{d_i}}Y_{hj}}) 
\end{equation}
Similarly, the weight of recovered interaction $Y^t_{ij}$ is computed as:
\begin{equation}
    W^t_{ij} = \exp(-\frac{\sum_{t_l \in {\cal N}^k_{t_j}}C^t_{lj}Y_{il}}{\sum_{t_l \in {\cal N}^k_{t_j}}Y_{il}}) 
\end{equation}
The higher the weight, the more reliable the recovered interaction.
By weighted aggregation of the $\bm{Y}_d$ and $\bm{Y}_t$, we obtain the final recovered interaction matrix $\bm{Y}^{dt} \in \mathbb{R}^{n \times m}$:
\begin{equation}
Y^{dt}_{ij} = \frac{W^d_{ij}Y^{d}_{ij}+W^t_{ij}Y^{t}_{ij}}{W^d_{ij}+W^t_{ij}}, \ i=1,\cdots,n; j=1,\cdots,m
\label{eq:Ydt}
\end{equation}
By multiplying with the local imbalance-based weight, interactions recovered by more reliable similarities have more influence on $\bm{Y}^{dt}$.
The values in $\bm{Y}^{dt}$ are in the range of [0,1]. Furthermore, because ``1"s in $\bm{Y}$ denotes reliable interactions that do not need any update, a correction process is applied to the reconstructed interaction matrix to ensure the consistency of known interactions:
\begin{equation}
\bm{Y}^{dt} = \max\{\bm{Y}^{dt}, \bm{Y}\}
\label{eq:Ydt2}
\end{equation}
where $\max$ is the element wise maximum operator.

In the prediction phase, the estimated interaction between drug $d_u$ and target $t_v$ is calculated as: 
\begin{equation}
\hat{Y}_{uv} = \left\{
\begin{aligned}
& \frac{1}{z_{d_u}}\sum_{d_i \in {\cal N}^k_{d_u}}\eta^{i'-1}\bar{S}^d_{ui}Y^{dt}_{iv} , \text{ if } d_u \notin D \text{ and } t_v \in T \\
& \frac{1}{z_{t_v}}\sum_{t_j \in {\cal N}^k_{t_v}}\eta^{j'-1}\bar{S}^t_{vj}Y^{dt}_{uj} , \text{ if } d_u \in D \text{ and } t_v \notin T \\
& \frac{1}{z_{uv}}\sum_{(d_i,t_j) \in {\cal N}^k_{d_ut_v}} \eta^{i'+j'-2}({\bar{S}^d_{ui}})^{r_d}({\bar{S}^t_{vj}})^{r_t}Y^{dt}_{ij} \\&\qquad  \text{if } d_u \notin D \text{ and } t_v \notin T 
\end{aligned}
\right.
\label{eq:WkNNIR}
\end{equation}
where $r_d =\min\{1, L^d_{uv}/L^t_{uv}\} $ and $r_t= \min\{1, L^t_{uv}/L^d_{uv}\}$ are the coefficients controlling the weights of drug similarities and target similarities respectively. The smaller  $r_d$ is, the larger drug similarities become, as both similarities and $r_d$ are between [0,1]. $L^d_{uv} = \sum_{(d_i,t_j) \in {\cal N}^k_{d_ut_v}} C^d_{ij}Y_{ij}$ and $L^t_{uv} = \sum_{(d_i,t_j) \in {\cal N}^k_{d_ut_v}} C^t_{ij}Y_{ij}$ are the sum of drug-based and target-based local imbalance of neighbor pairs of $(d_u, t_v$) respectively.
Compared with Eq.\eqref{eq:wknn23} and \eqref{eq:wknn4} in WkNN, there are two improvements made in WkNNIR. The first one is the utilization of recovered interaction matrix with more sufficient interaction information. 
The second advantage of WkNNIR is the more reliable similarity assessment via local imbalance by using $r_d$ and $r_t$ in S4. 

\begin{figure}[htp]
\centering
\includegraphics[width=0.6\textwidth]{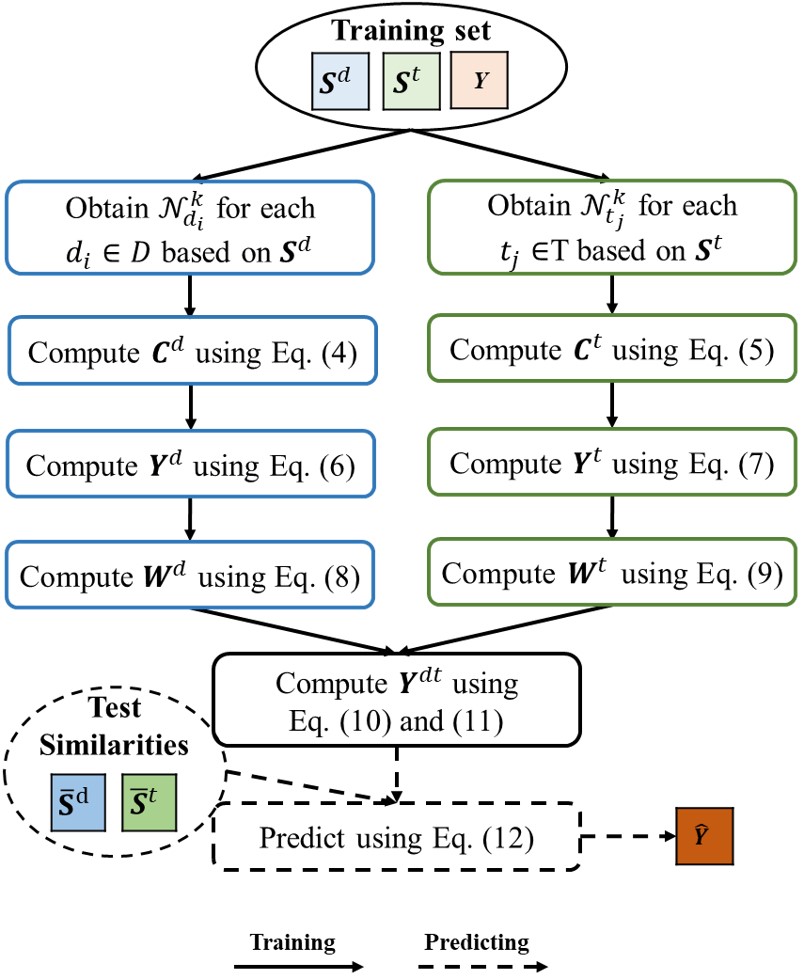}
\caption{The workflow of WkNNIR. Solid and dashed arrows denote steps in the training and prediction phases, respectively. } 
\label{fig:WkNNIR_workflow}
\end{figure}

The workflow of WkNNIR is presented in Fig. \ref{fig:WkNNIR_workflow}. In the training phase, it sequentially computes ${\cal N}^k_{d_i}$ for each training drug $d_i \in D$, drug-based local imbalance matrix $\bm{C}^d$, drug-based recovery interaction matrix $\bm{Y}^d$ and $\bm{W}^d$ representing weights of $\bm{Y}^d$, using the training set. Similar variables relating to targets, namely ${\cal N}^k_{t_j}$, $\bm{C}^t$, $\bm{Y}^t$ and $\bm{W}^t$, are calculated too. Then, the final interaction matrix $\bm{Y}^{dt}$ incorporating the recovered interactions in both $\bm{Y}^d$ and $\bm{Y}^t$ is obtained according to Eq.\eqref{eq:Ydt} and Eq.\eqref{eq:Ydt2}. In the prediction phase, given test drug and/or target similarities, the estimated interaction matrix $\hat{Y}$ is obtained based on the recovered interaction matrix $\bm{Y}^{dt}$ according to Eq.\eqref{eq:WkNNIR}.

\section{Ensembles of DTI Prediction Models}
Ensemble methods integrate multiple models that solve the same task, and therefore reduce the generalization error of single models. 
In this section, we propose three ensemble methods, namely ERS, EGS and ELS, which employ random sampling, global imbalance based sampling, and local imbalance based sampling respectively. The three proposed ensemble methods follow the same framework and can be applied to any DTI prediction method to further improve its accuracy.
We firstly introduce the ensemble framework and then describe the sampling strategies used in each method.

\subsection{Ensemble Framework}
The proposed ensemble framework learns multiple models based on diverse sampled training subsets. To adapt it to the DTI prediction task, the proposed framework needs to be modified in the following aspects: adjusting the sampled training subsets generation in the training phase as well as adding dynamic ensemble selection and similarity vector projection processes in the prediction phase. 

The pseudo code of training an ensemble model is shown in Algorithm \ref{al:Train_E}. 
In the training phase, the sampling probabilities for drugs and targets, denoted as $\bm{p}^d \in \mathbb{R}^n$ and $\bm{p}^t \in \mathbb{R}^m$, are initially computed (Algorithm \ref{al:Train_E}, line 2-3), where $\sum_{i=1}^n p^d_i =1$ and $\sum_{j=1}^m p^t_j =1$. The calculation of the sampling probabilities is different in each of the three methods and will be illustrated in Section \ref{sec:sampling_probability}. 
Then, $q$ base models are trained iteratively. 
For the $i$-th base model, the $nR$ sized drug subset $D^i$ is sampled from $D$ without replacement according to $\bm{p}^d$ (Algorithm \ref{al:Train_E}, line 5), i.e. drugs with larger sampling probability have a greater chance to be added to $D^i$, where $R$ is a user-specified sampling ratio controlling the number of selected drugs.
In a similar way, the $mR$ sized target subset $T^i$ is derived from $T$ based on $\bm{p}^t$ (Algorithm \ref{al:Train_E}, line 6). 
In the next step, based on the $D^i$ and $T^i$, we form a training subset which consists of a drug similarity sub-matrix $\bm{S}^{di}$ retaining similarities between drugs in $D^i$, a target similarity sub-matrix $\bm{S}^{ti}$ preserving the similarities between targets in $T^i$ and an interaction sub-matrix $\bm{Y}^i$ storing interactions involving both $D^i$ and $T^i$ (Algorithm \ref{al:Train_E}, line 7). 
Finally, the base model $M_i$ is trained on the obtained training subset (Algorithm \ref{al:Train_E}, line 8).

\begin{algorithm}
 \SetKwData{Left}{left}\SetKwData{This}{this}\SetKwData{Up}{up}
 \SetKwInOut{Input}{input}\SetKwInOut{Output}{output}
 
 \Input{training drug set: $D$, training target set: $T$, drug similarity matrix: $\bm{S}^d$, target similarity matrix: $\bm{S}^t$, interaction matrix: $\bm{Y}$, sampling ratio: $R$, ensemble size: $q$}
 \Output{ensemble model: $M$,  drug subsets: $D'$, target subsets: $T'$}
 $M, D', T' \leftarrow \emptyset$ \;
 Calculate the sampling probability for drugs $\bm{p}^d$\;
 Calculate the sampling probability for targets $\bm{p}^t$\;
\For{$i \leftarrow 1$ \KwTo $q$}{
    $D^i \leftarrow$ sample $nR$ drugs from $D$ based on $\bm{p}^d$ \;
    $T^i \leftarrow$ sample $mR$ targets from $T$ based on $\bm{p}^t$ \;
    Construct training subset $\bm{S}^{di}, \bm{S}^{ti}, \bm{Y}^i$ based on $D^i$ and $T^i$ \;
    $M_i \leftarrow$ Train($\bm{S}^{di}, \bm{S}^{ti}, \bm{Y}^i, D^i, T^i$) \;
    $M \leftarrow M \cup M_i$ \; 
    $D' \leftarrow D' \cup D^i$ \;
    $T' \leftarrow T' \cup T^i$ \;
}
\KwRet{$M, D', T'$} \;
 \caption{Training Ensembles of DTI Prediction Models}
  \label{al:Train_E}
\end{algorithm}

The prediction process of the ensemble framework is illustrated in Algorithm \ref{al:Predict_E}. The input similarity $\bm{s}^d_u$ ($\bm{s}^t_v$) is equivalent to $\bm{S}^d_{u\cdot}$ ($\bm{S}^t_{v\cdot}$) if $d_u \in D$ ($t_v \in T$), and $\bar{\bm{S}}^d_{u\cdot}$ ($\bar{\bm{S}}^t_{v\cdot}$) otherwise. 
Dynamic ensemble selection takes place first if the test drug-target pair ($d_u,t_v$) follows the condition of S2 or S3. 
Specifically, when ($d_u,t_v$) follows S2, $M_i$ is discarded if test target $t_v \notin T_i$ (Algorithm \ref{al:Predict_E}, line 2-4). Compared with other base models involving $t_v$ in the training process, $M_i$ misses the information of interaction regarding $t_v$ and usually leads to a lower prediction accuracy for ($d_u,t_v$).
Analogously, the models whose corresponding training drug subset does not contain $d_u$ are discarded if ($d_u,t_v$) follows S3 (Algorithm \ref{al:Predict_E}, line 5-7). 
Dynamic ensemble selection is not applied to S4 ($d_u \notin D$ and $t_v \notin T$), because both $d_u$ and $t_v$ are new emerging drugs and targets for all base models in S4. 
In the next steps, all retained base models give their prediction, which are eventually averaged to obtain the final predicted score $\hat{Y}_{uv}$ (Algorithm \ref{al:Predict_E}, line 9-14).
It should be noticed that there are two projection steps to ensure the similarity vectors of the drug and target in the test pair fit the input of each base model (Algorithm \ref{al:Predict_E}, line 10-11). 
As each base model is trained based on a subset of drugs and targets, the similarity vector for $d_u$ and $t_v$ should be projected to the low dimensional space characterized by the drug and target subset used in the corresponding base model. Specifically, the projection of $\bm{s}^d_u$ on $D_i$ maintains the similarities between $d_u$ and drugs in $D_i$ and deletes other elements in $\bm{s}^d_u$. For example, given a similarity vector $[0.1,0.2,0.3,0.4,0.5]$, its projection on drug subset $\{d_1, d_2, d_4\}$ is $[0.1,0.2,0.4]$.

\begin{algorithm}
 \SetKwData{Left}{left}\SetKwData{This}{this}\SetKwData{Up}{up}
 \SetKwInOut{Input}{input}\SetKwInOut{Output}{output}
 
 \Input{ensemble of DTI prediction model: $M$, drug subsets: $D'$, target subsets: $T'$, training drug set: $D$, training target set: $T$, test drug-target pair: ($d_u$,$t_v$), similarity of $d_u$ to $D$: $\bm{s}^d_u$, similarity of $t_v$ to $T$: $\bm{s}^t_v$}
 \Output{predicting interaction between $d_u$ and $t_v$: $\hat{Y}_{uv}$}

\For(\tcc*[h]{Dynamic ensemble selection}){$i \leftarrow 1$ \KwTo $|M|$}{
    \If(\tcc*[h]{S2}){$d_u \notin D$ and $t_v \in T$}{
        \If{$t_v \notin T_i$}{
            $M \leftarrow M - M_i$ \;
        }
     }
    \If(\tcc*[h]{S3}){$d_u \in D$ and $t_v \notin T$}{
        \If{$d_u \notin D_i$}{
            $M \leftarrow M - M_i$ \;
        }
    }
}
$\hat{Y}_{uv} \leftarrow 0$ \;
\For(\tcc*[h]{Predicting}){$i \leftarrow 1$ \KwTo $|M|$}{
    $\bm{s}'^{d}_u \leftarrow$ Project $\bm{s}^{d}_u$ on $D^i$ \;   
    $\bm{s}'^{t}_v \leftarrow$ Project $\bm{s}^{t}_v$ on $T^i$ \;   
    $Y'_{uv} \leftarrow \text{Predict}(M_i, \bm{s}'^{d}_u, \bm{s}'^{t}_v, d_u, t_v)$ \;
    $\hat{Y}_{uv} \leftarrow \hat{Y}_{uv}+Y'_{uv}$ \;
}
$\hat{Y}_{uv} \leftarrow \hat{Y}_{uv}/|M|$ \;

\KwRet{$\hat{Y}_{uv}$} \;
 \caption{Predicting of Ensemble DTI Prediction}
  \label{al:Predict_E}
\end{algorithm}

\subsection{Sampling Probability \label{sec:sampling_probability}}
Sampling probabilities determine the opportunity of each drug and target being used to train base models, which play a key role in the proposed ensemble framework. As we mentioned before, the three proposed ensemble methods employ different sampling probabilities. 

ERS adopts the sampling probabilities following the uniform distribution, i.e. $p^d_i = 1/n$ and $p^t_j = 1/m$, where $i=1,2,...n$ and $j=1,2,...m$. In this way, each drug and target has an equal chance to be selected. 

In DTI data, interacting drug-target pairs are heavily outnumbered by non-interacting ones, resulting in an imbalanced distribution within the global interaction matrix.
To relieve this global imbalance, EGS forms training subsets by biasing the sampling process to include drugs and targets having more interactions. 
Moreover, another reason for emphasis on drugs and targets with dense interactions is that they are more informative than others with fewer interactions. 
In EGS, the sampling probability of each drug (target) is proportional to the number of its interactions:
\begin{equation}
\begin{aligned}
p^d_i = \frac{\sigma+\sum_{j=1}^m Y_{ij}}{n\sigma+\sum_{h=1}^n \sum_{j=1}^m Y_{hj}} , \quad i=1,2,...n \\
p^t_j =\frac{\sigma+\sum_{i=1}^n Y_{ij}}{m\sigma+\sum_{i=1}^n \sum_{h=1}^m Y_{ih}}, \quad j=1,2,...m
\end{aligned}
\label{eq:EGS}
\end{equation}
where $\sigma$ is a smoothing parameter. By using Eq. \eqref{eq:EGS}, the drugs and targets with more interactions are more likely to be selected in the sampling procedure.

Apart from the global imbalance, the local imbalance could also be used to assess the importance of that drug (target).
According to Eq.\eqref{eq:Cd_ij}, higher $C^d_{ij}$ means that $d_i$ is surrounded by more drugs that have opposite interactivity to $t_j$. In such cases, correctly predicting $Y_{ij}$ using drug similarities would be difficult. 
By accumulating the local imbalance (difficulty) of $d_i$ for all interacting targets, we arrive at the local imbalance based importance of $d_i$:
\begin{equation}
LI^d_i = \sum_{j=1}^m C^d_{ij}\llbracket Y_{ij}=1 \rrbracket
\label{eq:LIdi}
\end{equation}

$LI^d_i$ is a weighted sum version of $\sum_{j=1}^m Y_{ij}$, where the interaction which is difficult to be predicted correctly is assigned a higher weight. Compared with EGS only counting the interactions of drugs, Eq. \eqref{eq:LIdi} emphasizes on drugs having more difficult interactions. 

Similarly, the local imbalance based importance of $t_j$ is defined as: 
\begin{equation}
LI^t_j = \sum_{i=1}^n C^t_{ij}\llbracket Y_{ij}=1 \rrbracket     
\end{equation}
Based on the definition of local imbalanced based importance, the key idea in ELS is that it encourages more \textit{difficult} drugs and targets to be learned by more base models, reducing thereby the corresponding error to the greatest extent possible.
In ELS, the sampling probability is proportional to the corresponding local imbalance based importance:
\begin{equation}
\begin{aligned}
p^d_i = \frac{\sigma+LI^d_i}{n\sigma+\sum_{h=1}^n LI^d_h} , \quad i=1,2,...n \\
p^t_j =\frac{\sigma+LI^t_j}{m\sigma+\sum_{h=1}^m LI^t_h}, \quad j=1,2,...m
\end{aligned}
\label{eq:ELS}
\end{equation}

\begin{figure}[htp]
\centering
\includegraphics[width=0.6\textwidth]{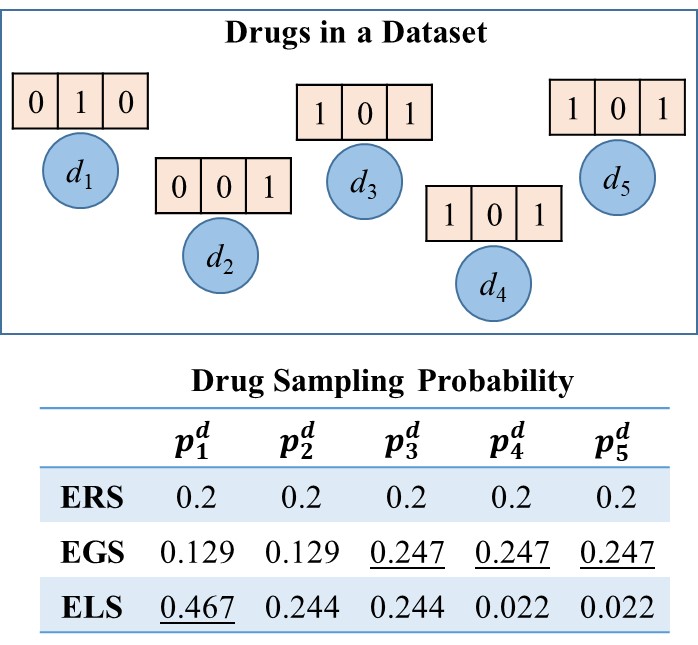}
\caption{The differences of the three sampling probabilities. The top half shows the drugs' location (blue circles) and their interactions (light red rectangles) of a dataset including five drugs and three targets. The table in the bottom half lists the drug sampling probabilities computed by the three ensemble methods on this dataset with $\sigma=0.1$ and $k=2$ } 
\label{fig:Sampling_Prob}
\end{figure}

In addition, we exemplify the differences of three sampling probabilities in Fig. \ref{fig:Sampling_Prob}. In ERS, all drugs have equal chance to be selected. EGS is more likely to choose $d_3, d_4 \text{ and } d_5$ that have more interacting targets. ELS will select $d_1$ with higher probability, because $d_1$ is near (similar) to drugs ($d_2$ and $d_3$) having different interactions and is therefore more difficult to be learned than other drugs.

\section{Experiments}
In this section, the datasets and the evaluation protocol used in the experiments are presented firstly. Then, the predictive performance and parameter analysis of proposed WkNNIR and three ensembles are reported. Finally, newly discovered interactions found by our methods are presented.

\subsection{Dataset}
Five benchmark DTI datasets are used in our empirical study. Four of them are gold standard datasets originally provided by \citep{Yamanishi2008PredictionSpaces}, of which each corresponds to a target protein family, namely Nuclear Receptors (NR), G-protein coupled receptors (GPCR), Ion Channel (IC) and Enzyme (E). The last dataset (DB), obtained from \citep{Kuang2015AnInteraction}, was derived from DrugBank \citep{Wishart2018DrugBank2018}. Table \ref{ta:Dataset} lists the information about the five datasets. Sparsity is the proportion of interacting drug-target pairs which indicates the global imbalance of the dataset. $LI^d$ and $LI^t$ are the drug-based and target-based local imbalance of the whole dataset respectively: 
\begin{equation}
LI^d = \frac{\sum_{i=1}^n \sum_{j=1}^m C^d_{ij}Y_{ij}}{\sum_{i=1}^n \sum_{j=1}^m Y_{ij}}
\label{eq:LId}
\end{equation}
\begin{equation}
LI^t = \frac{\sum_{i=1}^n \sum_{j=1}^m C^t_{ij}Y_{ij}}{\sum_{i=1}^n \sum_{j=1}^m Y_{ij}}
\label{eq:LIt}
\end{equation}
where $k$ is the number of neighbors, which is set to 5. Smaller $LI^d$ ($LI^t$) values indicate more reliable drug (target) similarities and easier S2 (S3) prediction task.


\begin{table}[htp]
\centering
\caption{Statistic of DTI datasets}
\label{ta:Dataset}
\begin{tabular}{@{}ccccccc@{}}
\toprule
Dataset & \# Drugs & \# Targets & \# Interactions & Sparsity & $LI^d$ & $LI^t$ \\ \midrule
NR & 54 & 26 & 90 & 0.064 & 0.658 & 0.764 \\
GPCR & 223 & 95 & 635 & 0.03 & 0.707 & 0.644 \\
IC & 210 & 204 & 1476 & 0.035 & 0.729 & 0.323 \\
E & 445 & 664 & 2926 & 0.01 & 0.737 & 0.35 \\ 
DB & 786 & 809 & 3681 & 0.006 & 0.653 & 0.528 \\
\bottomrule
\end{tabular}
\end{table}

\subsection{Evaluation protocol} \label{Evaluation_protocol}

Three types of cross validation (CV) are conducted to examine the prediction methods in three prediction settings, respectively. In S2, the drug wise CV is applied where one drug fold along with their corresponding rows in $\bm{Y}$ are separated for testing. In S3, the target wise CV is utilized where one target fold along with their corresponding column in $\bm{Y}$ are left out for testing. The block wise CV, which splits a drug fold and target fold along with the interactions between them (which is a sub-matrix of $\bm{Y}$) for testing and uses interactions between remaining drugs and targets for training, is applied to S4.
Two repetitions of 10-fold CV are applied to S2 and S3, and two repetitions of 3-fold block wise CV which contains 9 block folds generated by 3 drug folds and 3 target folds are applied to S4. 

The Area Under the Precision-Recall curve (AUPR) which heavily punishes highly ranked false positive predictions \citep{Schrynemackers2013OnNetworks} is used to evaluate the performance of inductive DTI prediction approaches in our experiments. In addition, the Wilcoxon signed rank test at 5\% level is utilized to examine the statistically significant difference between our methods and the compared ones. 

In the experiments, we firstly compare the proposed WkNNIR with six DTI prediction approaches, namely ALADIN \citep{Buza2017ALADIN:Prediction}, BICTR \citep{Pliakos2020Drug-targetReconstruction}, BLM-MLkNN \citep{Zhang2015PredictingLearning,Schrynemackers2015ClassifyingInference}, BLM-NII \citep{Mei2013Drug-targetNeighbors}, MLkNNSC \citep{Shi2015PredictingClustering}, NRLMF \citep{Liu2016NeighborhoodPrediction}, as well as WkNN which is the baseline of WkNNIR without interaction recovery and local imbalance weighting. All comparing methods except for BICTR either follow the transductive DTI prediction task or work for partial prediction settings. Therefore, we extend those methods to handle the inductive DTI prediction problem and all prediction settings as follows:
\begin{itemize}
    \item ALADIN \citep{Buza2017ALADIN:Prediction}: ALADIN works for S2 and S3 within the inductive scheme. As a BLM approach, ALADIN could deal with S4 by using the two-step learning strategy.
    \item BLM-MLkNN \citep{Zhang2015PredictingLearning,Schrynemackers2015ClassifyingInference}:  An individual MLkNN model could only deal with S2 and S3 following the inductive scheme. To deal with S4, the MLkNN is embedded in the BLM and the two-step learning strategy is adopted.
    \item MLkNNSC \citep{Shi2015PredictingClustering}:  MLkNNSC is initially proposed to predict interactions for test drugs (S2) and test drug-target pairs (S4) and it could work as an inductive approach directly. To handle S3 prediction setting, MLkNNSC is extended via applying clustering for drugs to obtain super-drugs and training MLkNN models for drugs and super-drugs, respectively. 
    \item BLM-NII \citep{Mei2013Drug-targetNeighbors}: BLM-NNI is a transductive method and able to tackle all prediction settings. To enable BLM-NNI to adapt the inductive DTI prediction, we modify its training and prediction process by confining that similarities for test drugs (targets) are available in prediction phase only.
    \item NRLMF \citep{Liu2016NeighborhoodPrediction}: NRLMF predicts interactions for S2, S3, and S4 in the transductive way. Similar to BLM-NII, NRLMF is altered to an inductive method by excluding similarities for test drugs and targets from the input of the training phase.
\end{itemize}
Moreover, BICTR trains an ensemble of bi-clustering trees on a reconstructed interaction matrix which is completed by NRLMF, and the input of the tree-ensemble model is the drug and target features. Hence, the similarities are utilized as features in BICTR, i.e. the feature vector of a drug (target) is its similarities to all training drugs (targets).

When it comes to ensemble methods i.e. ERS, EGS and ELS, four comparing DTI methods, namely ALADIN, BLM-NII, MLkNNSC, NRLMF and the two proposed neighborhood based approaches (WkNN and WkNNIR) are utilized as base model. We do not employ BICTR and BLM-MLkNN as base model, because the former one is already an ensemble model and the latter one is the worst method in most cases.

The parameter settings of compared and proposed methods are listed in Table \ref{ta:Parameters}. The values for used parameters are selected by performing the inner CV on the training set. Specifically, the 5-fold inner CV is applied to S2 and S3, and the 2-fold inner CV is applied to S4 on GPCR, IC, E and DB datasets. For NR dataset containing fewer drugs and targets, splitting NR into fewer folds generates small-sized training sets that may vary the distribution of the whole dataset during the inner CV procedure. This leads to the unreliability of chosen optimal parameter settings. To avoid this issue, we apply the 10-fold inner CV for S2 and S3 and the 3-fold inner CV for S4 on NR dataset.



\begin{table}[htp]
\centering
\caption{Parameter settings}
\label{ta:Parameters}
\begin{tabular}{@{}cc@{}}
\toprule
Method & Values or ranges of parameters \\ \midrule
ALADIN & \#models=25,  $k \in$\{1,2,3,5,7,9\}, \#features$\in$\{10,20,50\}\\
BICTR & \#trees=100, minimal \# samples in leaf=1 (2) in S3 and S4 (S2) \\
BLM-MLkNN &  $k \in$\{1,2,3,5,7,9\} \\
BLM-NII & $\gamma$ = 1, $\alpha \in$  \{0, 0.1,...,1.0\},  $\lambda \in\{2^{-5},2^{-4},...,2^{0}\}$ \\
MLkNNSC  & cut-off threshold= 1.1, $k \in$\{1,2,3,5,7,9\}\\
NRLMF & $c$=5, $k$=5, $r$ $\in$ \{50,100\}, $\lambda_t, \lambda_d, \alpha, \beta \in\{2^{-5},2^{-4},...,2^{0}\}$\\ 
WkNN  & $k \in$\{1,2,3,5,7,9\}, $\eta \in$  \{0.1,0.2,...,1.0\}\\
WkNNIR  & $k \in$\{1,2,3,5,7,9\}, $\eta \in$  \{0.1,0.2,...,1.0\}\\ \midrule
ERS & $q=30$, $R=0.95$  \\
EGS & $q=30$, $R=0.95$  $\sigma=1.0 (0.1)$ for DB dataset (others)\\
ELS & $q=30$, $R=0.95$, $\sigma=1.0 (0.1)$ for DB dataset (others), $k=5$ \\
 \bottomrule
\end{tabular}
\end{table}

\subsection{Results}
In this part, the obtained results comparing the proposed WkNNIR and its baseline WkNN to other competitors are presented and discussed. Next, the results of the three proposed ensemble methods with six different base models are reported. 

Table \ref{ta:DTImodel_result} shows the AUPR results for the compared approaches in various prediction prediction settings, where ``*" following the numerical results indicates that the corresponding method is statistically different from WkNNIR using Wilcoxon signed rank test at 5\% level.
WkNNIR achieves the best average rank in all settings. It is significantly superior to other methods in 83/105 cases and does not suffer any significant losses from other competitors. In addition, WkNNIR significantly outperforms WkNN in 13/15 cases, demonstrating the effectiveness of the utilization of the interaction recovery and local imbalance-based weights.
Then, we investigate the results in each prediction setting.
In S2, WkNNIR is the best method on all datasets. WkNN is the second best method and NRLMF comes next in most cases. 
In S3, for more difficult datasets (GPCR and DB having higher target-based local imbalance), WkNNIR is the top method, while for easier datasets (IC and E having lower target-based local imbalance) and the small-sized dataset (NR), WkNNIR achieves comparable performance to the corresponding best method without statistically significant difference. 
In S4, WkNNIR outperforms other competitors on the first three datasets and is slightly inferior to WkNN on E as well as BICTR on DB without significant difference. WkNN outperforms all other six competitors on the first four datasets, which indicates the effectiveness of our proposed neighbor pair based prediction function for S4.
Overall, WkNNIR surpasses the compared methods in S2 and S4 and is comparable with state-of-the-art approaches in S3.

\begin{sidewaystable}
\centering
\caption{Results of comparison inductive DTI prediction methods in terms of AUPR. The parenthesis is the rank of each method among all competitors. 
}
\label{ta:DTImodel_result}
\begin{tabular}{@{}cccccccccc@{}}
\toprule
Setting & Dataset & ALADIN & BLMNII & BLM-MLkNN & BICTR & MLkNNSC & NRLMF & WkNN & WkNNIR \\ \midrule
\multirow{6}{*}{S2} & NR & 0.433*(8) & 0.441*(6) & 0.436*(7) & 0.462*(4) & 0.456*(5) & 0.513(2) & 0.51*(3) & \textbf{0.539(1)} \\
 & GPCR & 0.306*(8) & 0.342*(4.5) & 0.326*(7) & 0.328*(6) & 0.342*(4.5) & 0.345*(3) & 0.369*(2) & \textbf{0.384(1)} \\
 & IC & 0.35(4) & 0.317*(6) & 0.298*(8) & 0.359(2) & 0.312*(7) & 0.343*(5) & 0.354*(3) & \textbf{0.363(1)} \\
 & E & 0.289*(7) & 0.26*(8) & 0.353*(3) & 0.338*(6) & 0.34*(5) & 0.352*(4) & 0.385*(2) & \textbf{0.396(1)} \\
 & DB & 0.41*(4) & 0.202*(8) & 0.365*(7) & 0.423(2) & 0.366*(6) & 0.386*(5) & 0.413*(3) & \textbf{0.425(1)} \\
 & \textit{AveRank} & 6.2 & 6.5 & 6.4 & 4 & 5.5 & 3.8 & 2.6 & \textbf{1} \\ \midrule
\multirow{6}{*}{S3} & NR & 0.383*(6) & 0.447*(3) & 0.37*(8) & 0.392*(5) & 0.38*(7) & \textbf{0.471(1)} & 0.443*(4) & 0.461(2) \\
 & GPCR & 0.517*(5) & 0.476*(8) & 0.493*(7) & 0.539*(3) & 0.511*(6) & 0.518*(4) & 0.541*(2) & \textbf{0.577(1)} \\
 & IC & \textbf{0.803(1)} & 0.787(6) & 0.777*(8) & 0.799(2) & 0.784(7) & 0.798(3.5) & 0.789*(5) & 0.798(3.5) \\
 & E & 0.758*(6) & 0.77*(4.5) & 0.75*(7) & 0.77*(4.5) & 0.748*(8) & \textbf{0.786(1)} & 0.776*(3) & 0.78(2) \\
 & DB & 0.569*(5) & 0.433*(8) & 0.563*(6) & 0.554*(7) & 0.579*(4) & 0.585(2) & 0.581*(3) & \textbf{0.595(1)} \\
 & \textit{AveRank} & 4.6 & 5.9 & 7.2 & 4.3 & 6.4 & 2.3 & 3.4 & \textbf{1.9} \\ \midrule
\multirow{6}{*}{S4} & NR & 0.095*(7) & 0.135(5) & 0.07*(8) & 0.154(3) & 0.117*(6) & 0.142(4) & 0.159(2) & \textbf{0.165(1)} \\
 & GPCR & 0.114*(6) & 0.121*(5) & 0.091*(8) & 0.135*(3) & 0.1*(7) & 0.134*(4) & 0.149*(2) & \textbf{0.158(1)} \\
 & IC & 0.206*(5) & 0.176*(6) & 0.131*(8) & 0.213(4) & 0.148*(7) & 0.215(3) & 0.216*(2) & \textbf{0.226(1)} \\
 & E & 0.128*(8) & 0.147*(5.5) & 0.147*(5.5) & 0.177*(4) & 0.146*(7) & 0.198(3) & \textbf{0.208(1)} & 0.202(2) \\
 & DB & 0.248(3) & 0.064*(8) & 0.183*(7) & \textbf{0.257(1)} & 0.205*(6) & 0.226*(5) & 0.247*(4) & 0.251(2) \\
 & \textit{AveRank} & 5.8 & 5.9 & 7.3 & 3 & 6.6 & 3.8 & 2.2 & \textbf{1.4} \\ \bottomrule
\end{tabular}
\end{sidewaystable}

As is deducted from the obtained results, the three prediction settings, namely S2, S3, and S4, are not equally challenging. 
In Table \ref{ta:DTImodel_result}, AUPR results of all methods in S4 are immensely lower than that in S2 and S3 on all datasets. This shows that predicting interactions between test drugs and test targets is the most challenging task. 
Comparing results in S2 and S3, we find that all methods achieve higher AUPR in S3 than S2 on GPCR, IC, E and DB whose $LI^t$ is lower than $LI^d$. While the performance of all methods, except for BLM-NII, in S3 is inferior to S2 on NR, whose $LI^t$ is higher than $LI^d$. This indicates that the difficulty of S2 and S3 could be estimated by comparing $LI^d$ and $LI^t$, e.g. S2 is harder than S3 if $LI^d$ is higher than $LI^t$, and vice versa. This also verifies the effectiveness of the local imbalance to assess the reliability of drug and target similarities.

As we previously stated, DTI datasets usually contain many missing interactions, e.g. the four gold standard datasets only contain interactions discovered before they were constructed (in 2007). To test the effectiveness of  WkNNIR on datasets with fewer missing interactions, we follow the procedure described in \cite{Peska2017Drug-targetApproach} to build updated gold standard datasets that include more validated interactions. Specifically, we add newly discovered interactions between drugs and targets in the original datasets recorded in the up-to-date version of KEGG \citep{Kanehisa2017KEGG:Drugs}, DrugBank \citep{Wishart2018DrugBank2018}, ChEMBL \citep{Mendez2019ChEMBL:Data} and Matador \cite{Gunther2008SuperTargetRelationships} databases. There are 175, 1350, 3201, 4640 interactions in the four updated dataset, denoted as NR1, GPCR1, IC1 and E1 respectively, with 85, 715, 1725 and 1714 new interactions appended.

Table \ref{ta:DTImodel_result_newdataset} lists the AUPR results of WkNNIR and four competitive comparing methods on the updated gold datasets. BLMNII, BLM-MLkNN and MLkNNSC are not included in the experiments on updated datasets due to their poor performance, as reported in Table \ref{ta:DTImodel_result}. In Table \ref{ta:DTImodel_result_newdataset}, we see that WkNNIR is still the best method in S2 and S3. In S4, WkNNIR is slightly inferior to WkNN, because the benefit of the interaction recovery used in WkNNIR is not significant when dealing with datasets having fewer missing interactions. In addition, comparing the results in Tables \ref{ta:DTImodel_result} and  \ref{ta:DTImodel_result_newdataset}, we find that the performance on the updated datasets with less missing interactions is usually better than in the original datasets. This verifies that missing interactions indeed hinder DTI prediction methods from achieving better performance.

\begin{table}[]
\caption{Results of comparing inductive DTI prediction methods on the updated gold standard datasets in terms of AUPR.}
\label{ta:DTImodel_result_newdataset}
\begin{tabular}{@{}ccccccc@{}}
\toprule
Setting & Dataset & ALADIN & BICTR & NRLMF & WkNN & WkNNIR \\ \midrule
\multirow{5}{*}{S2} & NR1 & 0.527(4) & 0.523(5) & 0.529(3) & 0.547(2) & \textbf{0.552(1)} \\
 & GCPR1 & 0.438(5) & 0.447(4) & 0.456(3) & 0.461(2) & \textbf{0.468(1)} \\
 & IC1 & 0.475(5) & 0.509(3) & 0.502(4) & 0.563(2) & \textbf{0.571(1)} \\
 & E1 & 0.301(5) & 0.331(4) & 0.362(3) & 0.368(2) & \textbf{0.382(1)} \\
 & \textit{AveRank} & 4.75 & 4 & 3.25 & 2 & \textbf{1} \\ \midrule
\multirow{5}{*}{S3} & NR1 & 0.465(5) & 0.518(4) & 0.52(3) & 0.542(2) & \textbf{0.565(1)} \\
 & GCPR1 & 0.874(2) & 0.871(3) & \textbf{0.888(1)} & 0.863(5) & 0.866(4) \\
 & IC1 & 0.75(2) & 0.748(3) & 0.731(5) & 0.747(4) & \textbf{0.762(1)} \\
 & E1 & 0.658(5) & 0.687(3) & \textbf{0.705(1)} & 0.685(4) & 0.695(2) \\
 & \textit{AveRank} & 3.5 & 3.25 & 2.5 & 3.75 & \textbf{2} \\ \midrule
\multirow{5}{*}{S4} & NR1 & 0.207(5) & \textbf{0.283(1)} & 0.27(3) & 0.271(2) & 0.259(4) \\
 & GCPR1 & 0.32(5) & 0.339(2) & 0.338(3) & 0.337(4) & \textbf{0.343(1)} \\
 & IC1 & 0.325(5) & 0.356(4) & 0.357(3) & \textbf{0.395(1)} & 0.388(2) \\
 & E1 & 0.146(5) & 0.181(4) & \textbf{0.214(1.5)} & \textbf{0.214(1.5)} & 0.209(3) \\
 & \textit{AveRank} & 5 & 2.75 & 2.63 & \textbf{2.13} & 2.5 \\ \bottomrule
\end{tabular}
\end{table}

The average rank of ensemble methods along with their embedded base models in terms of AUPR are summarized in Table \ref{ta:en_aveRank}.
The Base column denotes the average ranks of default base models, and the ``$\circ$" following the $AveRank$ denotes that the corresponding ensemble method is statistically superior to the base model using Wilcoxon signed rank test at 5\% level. The detailed numerical AUPR results are listed in Appendix Tables \ref{ta:en_S2}-\ref{ta:en_S4}.
We find that all three ensemble methods achieve better average rank compared to the base models in all prediction settings.
ELS is the most effective method and significantly outperforms the base models in all prediction settings. This is because ELS emphasizes on \textit{difficult} drugs and targets by considering local imbalance.
EGS aiming to reduce the global imbalance level comes next and its advantage over base models is significant as well.
ERS using a totally random sampling strategy is the third one and only achieves significant improvement in S3 and S4.

Furthermore, to check the effectiveness of the proposed ensemble methods on each base model, we calculate the average rank of the three ensemble methods on all datasets for each prediction setting and base model and pick up the best ones to show in Table \ref{ta:en_best_bm}. 
We divide the employed base models into two groups based on their performance in Table \ref{ta:DTImodel_result}: moderate base models (ALADIN, BLMNII and MLkNNSC) and good base models (NRLMF, WkNN, and WkNNIR).
Regarding the moderate base models, ERS, EGS and ELS are most effective on MLkNNSC, ALADIN and BLMNII, respectively.
When it comes to good models, ELS usually outperforms the other two ensemble methods. EGS is the top one only for NRLMF in S2 and WkNNIR in S3.
This suggests that ELS is more beneficial to base models with better prediction performance.

\begin{table}[]
\centering
\caption{The average ranks of ensemble methods over six base models and five datasets in three prediction settings.} 
\label{ta:en_aveRank}
\begin{tabular}{@{}ccccc@{}}
\toprule
Setting & Base & ERS & EGS & ELS \\ \midrule
S2 & 2.87 & 2.67 & 2.25$\circ$ & \textbf{2.22$\circ$} \\ 
S3 & 3.22 & 2.37$\circ$ & 2.22$\circ$ & \textbf{2.2$\circ$} \\
S4 & 3.72 & 2.43$\circ$ & 2.24$\circ$ & \textbf{1.59$\circ$} \\
\bottomrule
\end{tabular}
\end{table}

\begin{table}[]
\centering
\caption{The best ensemble method for each prediction setting and base model}
\label{ta:en_best_bm}
\begin{tabular}{@{}cccc@{}}
\toprule
Base Model & S2 & S3 & S4 \\ \midrule
ALADIN & EGS & EGS & EGS \\
BLMNII & ELS & ELS & ELS \\
MLkNNSC & ERS & ERS & ERS \\ \midrule
NRLMF & EGS & ELS & ELS \\
WkNN & ELS & ELS & ELS \\
WkNNIR & ELS & EGS & ELS \\ \bottomrule
\end{tabular}
\end{table}

\subsection{Parameter Analysis}
Here, we analyze the influence of parameter settings on WkNNIR and three ensemble methods.

\begin{figure*}[h]
\centering
\subfigure[Different settings of $k$]{\includegraphics[width=0.45\textwidth]{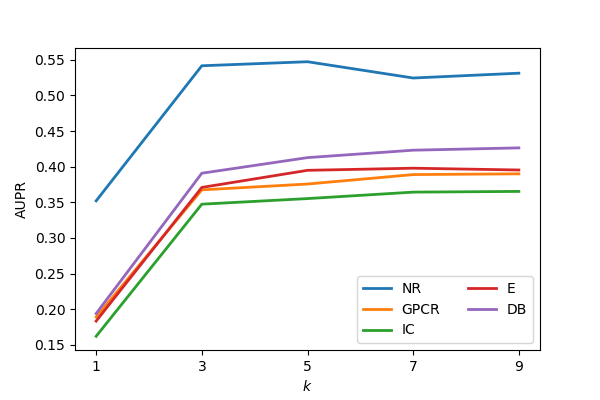}}
\subfigure[Different settings of $\eta$ ]{\includegraphics[width=0.45\textwidth]{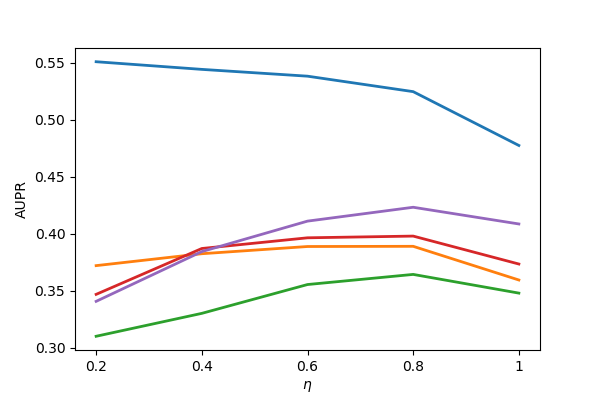}}
\caption{Performance of WkNNIR with different settings of $k$ and $\eta$ in S2 where $\eta$ is set as 0.8 for (a) and $k$ is set as 7 for (b)} 
\label{fig:VarParameters}
\end{figure*}

Firstly, we investigate the sensitivity of WkNNIR with respect to $k$ and $\eta$ in S2 which is shown in Fig.\ref{fig:VarParameters}. 
The performance of WkNNIR improves a lot from $k=1$ to $k=3$, as more neighbors are exploited. However, for values of $k$ larger than 3, the accuracy of WkNNIR plateaus, indicating that extra neighbors do not provide additional benefits.
In terms of $\eta$, median values (around 0.6 and 0.8) lead to the best performance on all datasets except for NR. The lower (higher) $\eta$ diminishes (promotes) the influence of the relatively remote neighbors on the prediction, leading to performance deterioration. In NR, which contains tens of drugs and targets, the performance drops as $\eta$ increases, indicating that lower $\eta$ is suitable for the small-sized dataset.   
Similar accuracy trends with respect to $k$ and $\eta$ are observed in S3 and S4 as well.

\begin{figure*}[h]
\centering
\subfigure[Different settings of $R$]{\includegraphics[width=0.45\textwidth]{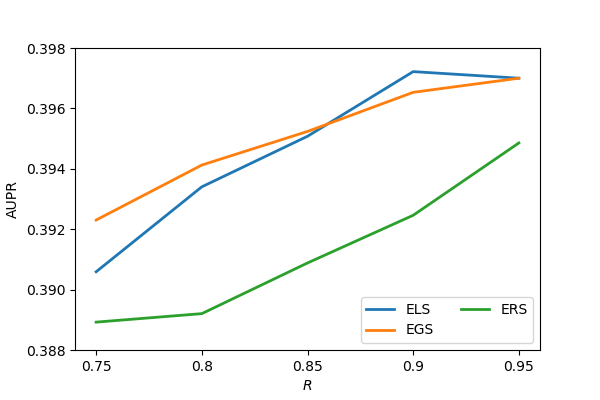}\label{fig:VP_en_R}}
\subfigure[Different settings of $q$]{\includegraphics[width=0.45\textwidth]{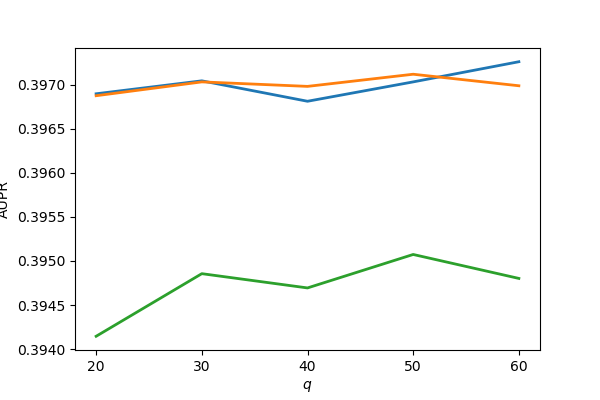}\label{fig:VP_en_q}} \\
\subfigure[Different settings of $\sigma$]{\includegraphics[width=0.45\textwidth]{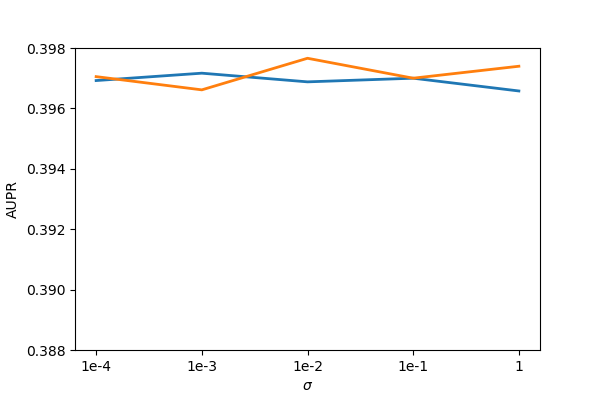}\label{fig:VP_en_sigma}}
\subfigure[Different settings of $k$]{\includegraphics[width=0.45\textwidth]{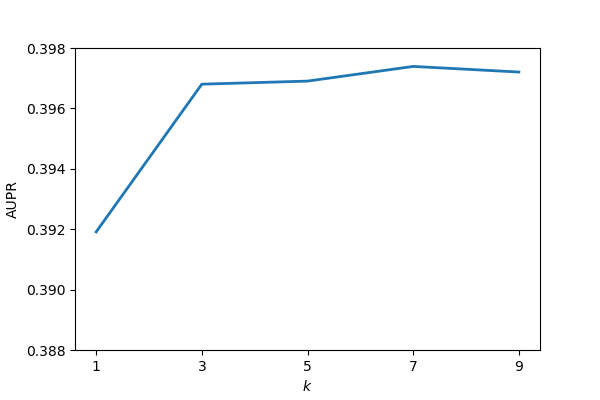}\label{fig:VP_en_k}}
\caption{Performance of ERS, EGS, ELS using WkNNIR as the base model under different parameter settings on E in S2} 
\label{fig:VarParameters_en}
\end{figure*}

Fig. \ref{fig:VarParameters_en} shows the performance of ensemble methods using WkNNIR as the base model under different parameter settings on E dataset in S2.
In Fig. \ref{fig:VP_en_R} concerning the sampling ratio, the performance of all three ensemble methods improve with the increase of $R$. This is mainly because selecting more drugs and targets contributes to training more accurate base models in them. 
Fig. \ref{fig:VP_en_q} presents the influence of ensemble size $q$ on each method. ERS achieves better performance when more base models are trained because larger $q$ would increase the chance of important drugs and targets being selected by ERS. Nevertheless, ERS training more than 50 base models cannot even surpass ELS and EGS with smaller ensemble size. EGS and ELS perform more stably under different ensemble sizes. This is because EGS and ELS guarantee that important drugs and targets would be more likely to be chosen and the benefit of training more base models is limited to them.
In Fig. \ref{fig:VP_en_sigma}, we find that EGS and ELS are insensitive to the smooth parameter $\sigma$. 
At last, as shown in Fig. \ref{fig:VP_en_k}, ELS reaches a plateau when $k$ which relates to the local imbalance of each drug and target is larger than 3.

\subsection{Discovery of New Interactions}
In this section, we examine the ability of WkNNIR and ensemble methods to discover new reliable DTIs in all prediction settings.
For each prediction setting and original gold standard dataset, we conducted cross-validation as illustrated in Section \ref{Evaluation_protocol} and combined the predictions of each fold to obtain the predicted scores of all drug-target pairs in the dataset.  
Then, we ranked all non-interacting drug-target pairs of the dataset according to their predicted scores and select the top 10 non-interacting pairs as the candidate newly predicted interactions. To verify the reliability of those new interactions, we checked whether they are included in the corresponding updated dataset that incorporate validated interactions from the last version of the KEGG, DrugBank, ChEMBL and Matador databases. 

The validated new interactions found by WkNNIR on the four original gold standard datasets are listed in Tables \ref{ta:new_dti_nr}-\ref{ta:new_dti_e}, where K, D, C, M indicate that the corresponding DTI is verified by KEGG, DrugBank, ChEMBL and Matador, respectively. WkNNIR could find at least one new DTI confirmed by external databases in each dataset and setting. 
Furthermore, we notice that there are many new DTIs discovered in only one prediction setting, e.g. for NR dataset D00690-hsa2098 is only found in S2 and D00954-hsa367 is only found in S4. 
Therefore, it is beneficial to find more new DTIs by examining the predictions from all settings. For example, four new DTIs could be found by looking at only setting S2 in the NR dataset. Nonetheless, nine new DTIs can be discovered if all settings are considered (D00348-hsa5915 is found in both S2 and S3).

In addition, Table \ref{tab:new_dti_number} summarizes the total number of validated new DTIs from top $X$ ($X \in \{10,20,30\}$) candidate pairs provided by WkNNIR and the other four competitors in all three prediction settings. The new DTIs discovered by more than two prediction settings are counted only once. It should be also mentioned that the maximum of new DTIs found from top $X$ candidates is $3X$, as we take three predictions settings into consideration. 
Regarding the top 10 candidate pairs, ALADIN finds the most new DTIs, followed by WkNNIR which discovers four fewer new DTIs among all datasets than ALADIN. However, picking only 10 candidate pairs may be not sufficient for practical applications, especially when a large amount of drug-target pairs need to be tested, e.g. E dataset contains 295,480 drug-target pairs.
When one increases the number of candidate pairs, WkNNIR becomes the most effective method detecting at least ten more new DTIs than other competitors. This indicates that WkNNIR outperforms other competitors in terms of the ability to predict new DTIs, particularly when many candidate pairs are considered.

Furthermore, in order to check whether we can find more new interactions by applying ensemble methods to WkNNIR, we collected validated new DTIs discovered by the three proposed ensemble methods using WkNNIR as the base model. 
Table \ref{ta:new_dti_en} lists the validated new DTIs only found by ensemble methods. 
ELS, EGS and ERS find 14, 9 and 12 new DTIs on the four datasets, respectively. 
This demonstrates that ensemble methods are indeed very promising in discovering new (not yet reported) DTIs.


\begin{table}[]
\centering
\caption{The validated new interactions predicted by WkNNIR in NR dataset}
\label{ta:new_dti_nr}
\begin{tabular}{@{}ccccc@{}}
\toprule
\multicolumn{1}{l}{Setting} & Drug & Target & Rank & Database \\ \midrule
\multirow{4}{*}{S2} & D00690 & hsa2908 & 3 & K, D \\
 & D00348 & hsa5915 & 5 & K \\
 & D00348 & hsa5916 & 6 & K, D \\
 & D05341 & hsa2099 & 10 & C \\ \midrule
\multirow{2}{*}{S3} & D00279 & hsa5468 & 8 & C \\
 & D00348 & hsa5915 & 9 & K \\ \midrule
\multirow{4}{*}{S4}  & D00954 & hsa367 & 1 & D \\
 & D00067 & hsa2100 & 3 & K \\
 & D00182 & hsa367 & 9 & D \\
 & D00554 & hsa2100 & 10 & K \\ \bottomrule
\end{tabular}
\end{table}

\begin{table}[]
\centering
\caption{The validated new interactions predicted by WkNNIR in GPCR dataset}
\label{ta:new_dti_gpcr}
\begin{tabular}{@{}ccccc@{}}
\toprule
Setting & Drug & Target & Rank & Database \\ \midrule
\multirow{8}{*}{S2} & D04625 & hsa154 & 1 & K, D, C \\
 & D02358 & hsa154 & 2 & D \\
 & D02147 & hsa153 & 3 & D, M \\
 & D00110 & hsa1128 & 4 & D \\
 & D02349 & hsa154 & 6 & K, D \\
 & D00095 & hsa155 & 7 & K \\
 & D00394 & hsa3269 & 8 & D \\
 & D00715 & hsa1129 & 10 & K, D \\ \midrule
\multirow{2}{*}{S3} & D00283 & hsa152 & 1 & K, D \\
 & D00283 & hsa1814 & 2 & D, M \\ \midrule
\multirow{4}{*}{S4} & D01103 & hsa1133 & 4 & K \\
 & D06056 & hsa3357 & 6 & D \\
 & D06056 & hsa3358 & 7 & D \\
 & D06056 & hsa3356 & 8 & D \\ \bottomrule
\end{tabular}
\end{table}

\begin{table}[]
\centering
\caption{The validated new interactions predicted by WkNNIR in IC dataset}
\label{ta:new_dti_ic}
\begin{tabular}{@{}ccccc@{}}
\toprule
Setting & Drug & Target & Rank & Database \\ \midrule
\multirow{5}{*}{S2} & D00438 & hsa779 & 1 & K, D \\
 & D00319 & hsa783 & 6 & D, M \\
 & D00319 & hsa786 & 7 & M \\
 & D00553 & hsa6329 & 9 & K \\
 & D00553 & hsa6334 & 10 & K \\ \midrule
\multirow{2}{*}{S3} & D00336 & hsa10060 & 1 & D \\
 & D00349 & hsa9254 & 9 & D \\ \midrule
\multirow{6}{*}{S4} & D00553 & hsa6328 & 4 & K \\
 & D00553 & hsa6334 & 5 & K \\
 & D01287 & hsa11280 & 6 & K \\
 & D04048 & hsa11280 & 7 & K \\
 & D01450 & hsa11280 & 8 & K \\ 
 & D00358 & hsa6332 & 9 & M \\\bottomrule
\end{tabular}
\end{table}

\begin{table}[]
\centering
\caption{The validated new interactions predicted by WkNNIR in E dataset}
\label{ta:new_dti_e}
\begin{tabular}{@{}ccccc@{}}
\toprule
Setting & Drug & Target & Rank & Database \\ \midrule
S2 & D00364 & hsa1565 & 10 & K, D \\ \midrule
\multirow{6}{*}{S3} & D00947 & hsa4129 & 1 & D \\
 & D00005 & hsa4128 & 2 & D \\
 & D05458 & hsa4128 & 3 & K, D \\
 & D01223 & hsa3988 & 5 & M \\
 & D00437 & hsa1585 & 6 & M \\
 & D00217 & hsa7173 & 10 & M \\
\midrule
\multirow{2}{*}{S4} & D00528 & hsa5150 & 3 & D \\
 & D00528 & hsa50940 & 4 & D \\ \bottomrule
\end{tabular}
\end{table}

\begin{table}[]
\centering
\caption{The number of validated new DTIs from top 10, 20 and 30 candidate pairs provided by DTI prediction methods}
\label{tab:new_dti_number}
\begin{tabular}{@{}ccccccc@{}}
\toprule
 & Dataset & ALADIN & BICTR & NRLMF & WkNN & WkNNIR \\ \midrule
\multirow{5}{*}{Top 10} & NR & 6 & 6 & 6 & 6 & \textbf{9} \\
 & GPCR & \textbf{18} & 15 & 13 & 14 & 14 \\
 & IC & \textbf{14} & 8 & 13 & 10 & 12 \\
 & E & 10 & \textbf{11} & 10 & 8 & 9 \\
 & \textit{Sum} & \textbf{48} & 40 & 42 & 38 & 44 \\ \midrule
{\multirow{5}{*}{Top 20}} & NR & 11 & 17 & 14 & 14 & \textbf{16} \\
 & GPCR & \textbf{33} & 31 & 22 & 22 & \textbf{33} \\
 & IC & \textbf{28} & 18 & \textbf{28} & 19 & 27 \\
 & E & 18 & 17 & 16 & 16 & \textbf{25} \\
 & \textit{Sum} & 90 & 83 & 80 & 71 & \textbf{101} \\ \midrule

\multirow{5}{*}{Top 30} & NR & 16 & 20 & 19 & 17 & \textbf{22} \\
 & GPCR & \textbf{47} & 41 & 33 & 37 & 42 \\
 & IC & 41 & 33 & \textbf{43} & 30 & 37 \\
 & E & 20 & 21 & 27 & 23 & \textbf{37} \\
 & \textit{Sum} & 124 & 115 & 122 & 107 & \textbf{138} \\ \bottomrule
\end{tabular}
\end{table}

\begin{table*}[]
\centering
\caption{The validated new interactions only found by ensemble methods with WkNNIR as base model}
\label{ta:new_dti_en}
\begin{tabular}{@{}cccccc@{}}
\toprule
Dataset & Setting & Drug & Target & Method (Rank) & Database \\ \midrule
\multirow{1}{*}{NR}  & S4 & D00950 & hsa367 & ERS(6), EGS(7), ELS(9) & D \\ \midrule
\multirow{3}{*}{GPCR} & \multirow{1}{*}{S2} & D00790 & hsa3269 & ERS(10) & D \\ \cmidrule(l){2-6}
 & \multirow{2}{*}{S4} & D00540 & hsa1131 & ERS(7), EGS(8), ELS(10) & K, D \\
 &  & D00394 & hsa1133 & ERS(8), EGS(10), ELS(9) & D \\ \midrule
\multirow{11}{*}{IC} & \multirow{10}{*}{S2} & D06172 & hsa6334 & ERS(6), ELS(6) & K \\
 &  & D06172 & hsa6329 & ERS(7), ELS(5) & K \\
 &  & D06172 & hsa6335 & ERS(8), ELS(8) & K \\
 &  & D06172 & hsa6323 & ERS(9), ELS(4) & K \\
 &  & D06172 & hsa6328 & ERS(10), ELS(7) & K \\
 &  & D00553 & hsa6335 & EGS(7), ELS(9) & K \\
 &  & D00553 & hsa6328 & EGS(8) & K \\
 &  & D00553 & hsa6323 & EGS(9) & K \\
 &  & D00553 & hsa6326 & EGS(10), ELS(10) & K \\
 &  & D06172 & hsa6326 & ELS(3) & K \\ \cmidrule(l){2-6}
 & S4 & D00438 & hsa779 & ERS(7), ELS(10) & K, D \\ \midrule
\multirow{4}{*}{E} & \multirow{4}{*}{S2} & D00560 & hsa1576 & ERS(9) & K, D, C \\
 &  & D00528 & hsa5150 & ERS(10) & D \\
 &  & D02441 & hsa762 & EGS(9), ELS(7) & K, D \\
 &  & D02441 & hsa760 & EGS(10), ELS(8) & K, D \\ \bottomrule
\end{tabular}
\end{table*}

\section{Conclusion}
In this paper, we propose a new neighborhood-based method (WkNNIR), which is able to deal with all types of interaction prediction settings and successful in handling missing interactions. Furthermore, we propose three ensemble methods, namely ERS, EGS and ELS, that integrate multiple DTI prediction models trained upon various sampled datasets to improve the performance of their embedded base model.
Both WkNNIR and ensemble methods were applied to five benchmark DTI prediction datasets.
The obtained results affirm the superiority of WkNNIR to other competing methods as well as its baseline WkNN. The performance improvement provided by the ensemble methods to six base models including WkNNIR is also verified. Particularly, ELS using local imbalance based sampling is the most outstanding ensemble approach. 
Subsequently, we demonstrated that our methods are able to predict reliable new drug–target interactions.

Although only chemical structure similarities and protein sequence similarities are used in this study, there are various kinds of similarities revealing the relationship between drugs (targets) in diverse aspects \citep{Li2019DrugEmbedding,Ding2020IdentificationFusion,Thafar2020DTiGEMS+:Techniques}. It is therefore desirable to investigate the effectiveness of our methods using different types of similarities as well as extending it to multi-modal or multi-view settings, where several types of similarities are integrated and exploited. 


\section*{Declarations}

\textbf{Funding}: Bin Liu is supported from the China Scholarship Council (CSC) under the Grant CSC No.201708500095. This research is also supported by the Flemish Government (AI Research Program).

\textbf{Conflict of interest}: The authors declare that they have no conflict of interest.

\textbf{Availability of data and code}: The datasets and code used in this paper are available at \href{{https://github.com/intelligence-csd-auth-gr/DTI_WkNNIR}}{\color{blue}{https://github.com/intelligence-csd-auth-gr/DTI\_WkNNIR}}.



\bibliographystyle{spbasic_nosort}      

\bibliography{LB}

\section*{Appendix}

\captionsetup[table]{name={Appendix Table},labelsep=colon}
\renewcommand{\thetable}{A\arabic{table}}
\setcounter{table}{0}
\appendix

\begin{table}[hb]
\centering
\caption{Results of ensemble methods along with their embedded base models in S2 in terms of AUPR. The parenthesis is the rank of each method among all competitors. 
}
\label{ta:en_S2}
\begin{tabular}{@{}cccccc@{}}
\toprule
Dataset & Base Model & Base & ERS & EGS & ELS \\ \midrule
\multirow{6}{*}{NR} & ALADIN & 0.433(3) & 0.433(3) & \textbf{0.446(1)} & 0.433(3) \\
 & BLMNII & 0.441(2) & 0.435(4) & 0.438(3) & \textbf{0.445(1)} \\
 & MLkNNSC & 0.456(4) & \textbf{0.495(1)} & 0.491(2) & 0.481(3) \\
 & NRLMF & \textbf{0.513(1)} & 0.496(4) & 0.506(2) & 0.503(3) \\
 & WkNN & 0.51(3.5) & 0.51(3.5) & 0.515(2) & \textbf{0.516(1)} \\
 & WkNNIR & 0.539(2) & 0.526(4) & \textbf{0.541(1)} & 0.534(3) \\ \midrule
 \multirow{6}{*}{GPCR} & ALADIN & 0.306(4) & \textbf{0.317(1)} & 0.313(2) & 0.311(3) \\
 & BLMNII & 0.342(2) & 0.341(3) & 0.34(4) & \textbf{0.344(1)} \\
 & MLkNNSC & 0.342(2.5) & \textbf{0.352(1)} & 0.342(2.5) & 0.341(4) \\
 & NRLMF & 0.345(4) & \textbf{0.358(1)} & 0.35(2.5) & 0.35(2.5) \\
 & WkNN & 0.369(2.5) & \textbf{0.37(1)} & 0.364(4) & 0.369(2.5) \\
 & WkNNIR & \textbf{0.384(1)} & 0.382(2) & 0.377(4) & 0.38(3) \\ \midrule
\multirow{6}{*}{IC} & ALADIN & \textbf{0.35(1)} & 0.34(4) & 0.346(2.5) & 0.346(2.5) \\
 & BLMNII & \textbf{0.317(1.5)} & 0.316(3.5) & 0.316(3.5) & \textbf{0.317(1.5)} \\
 & MLkNNSC & 0.312(4) & \textbf{0.321(1)} & 0.317(3) & 0.318(2) \\
 & NRLMF & 0.343(4) & \textbf{0.352(1)} & 0.347(3) & 0.351(2) \\
 & WkNN & 0.354(2) & 0.348(4) & \textbf{0.356(1)} & 0.351(3) \\
 & WkNNIR & \textbf{0.363(1)} & 0.359(4) & 0.361(2) & 0.36(3) \\ \midrule
\multirow{6}{*}{E} & ALADIN & 0.289(2.5) & 0.288(4) & \textbf{0.294(1)} & 0.289(2.5) \\
 & BLMNII & 0.26(4) & 0.262(3) & 0.264(2) & \textbf{0.273(1)} \\
 & MLkNNSC & 0.34(4) & \textbf{0.353(1)} & 0.348(2) & 0.344(3) \\
 & NRLMF & 0.352(4) & 0.365(2.5) & \textbf{0.367(1)} & 0.365(2.5) \\
 & WkNN & 0.385(3.5) & 0.385(3.5) & \textbf{0.388(1.5)} & \textbf{0.388(1.5)} \\
 & WkNNIR & 0.396(3) & 0.395(4) & \textbf{0.397(1.5)} & \textbf{0.397(1.5)} \\ \midrule
\multirow{6}{*}{DB} & ALADIN & 0.41(4) & 0.413(2) & 0.412(3) & \textbf{0.414(1)} \\
 & BLMNII & 0.202(4) & \textbf{0.211(1)} & 0.208(2) & 0.203(3) \\
 & MLkNNSC & 0.366(4) & \textbf{0.378(1)} & 0.372(3) & 0.374(2) \\
 & NRLMF & 0.386(3) & 0.375(4) & 0.393(2) & \textbf{0.394(1)} \\
 & WkNN & 0.413(3) & 0.394(4) & 0.414(1.5) & \textbf{0.414(1.5)} \\
 & WkNNIR & \textbf{0.425(2)} & 0.414(4) & \textbf{0.425(2)} & \textbf{0.425(2)} \\ \bottomrule
\end{tabular}
\end{table}

\begin{table}[]
\centering
\caption{Results of ensemble methods along with their embedded base models in S3 in terms of AUPR. The parenthesis is the rank of each method among all competitors. 
}
\label{ta:en_S3}
\begin{tabular}{@{}cccccc@{}}
\toprule
Dataset & Base Model & Base & ERS & EGS & ELS \\ \midrule
\multirow{6}{*}{NR} & ALADIN & 0.383(4) & \textbf{0.411(1)} & 0.403(2) & 0.393(3) \\
 & BLMNII & 0.447(3) & 0.445(4) & \textbf{0.454(1)} & 0.453(2) \\
 & MLkNNSC & 0.38(4) & \textbf{0.412(1)} & 0.405(2) & 0.402(3) \\
 & NRLMF & 0.471(3.5) & \textbf{0.48(1)} & 0.471(3.5) & 0.476(2) \\
 & WkNN & 0.443(4) & 0.444(3) & 0.45(2) & \textbf{0.451(1)} \\
 & WkNNIR & 0.461(4) & 0.466(3) & 0.477(2) & \textbf{0.479(1)} \\ \midrule
\multirow{6}{*}{GPCR} & ALADIN & \textbf{0.517(1)} & 0.511(3.5) & 0.513(2) & 0.511(3.5) \\
 & BLMNII & \textbf{0.476(1)} & 0.468(4) & 0.473(2) & 0.472(3) \\
 & MLkNNSC & 0.511(4) & \textbf{0.525(1)} & 0.515(2) & 0.514(3) \\
 & NRLMF & 0.518(3.5) & 0.518(3.5) & \textbf{0.527(1.5)} & \textbf{0.527(1.5)} \\
 & WkNN & 0.541(2) & \textbf{0.543(1)} & 0.539(4) & 0.54(3) \\
 & WkNNIR & 0.577(2) & 0.571(4) & 0.574(3) & \textbf{0.579(1)} \\ \midrule
\multirow{6}{*}{IC} & ALADIN & \textbf{0.803(2)} & 0.802(4) & \textbf{0.803(2)} & \textbf{0.803(2)} \\
 & BLMNII & \textbf{0.787(2)} & \textbf{0.787(2)} & 0.785(4) & \textbf{0.787(2)} \\
 & MLkNNSC & 0.784(4) & 0.791(3) & \textbf{0.795(1)} & 0.792(2) \\
 & NRLMF & 0.798(4) & 0.802(2) & 0.801(3) & \textbf{0.806(1)} \\
 & WkNN & 0.789(4) & 0.792(3) & 0.793(2) & \textbf{0.794(1)} \\
 & WkNNIR & 0.798(4) & 0.799(3) & \textbf{0.802(1)} & 0.8(2) \\ \midrule
\multirow{6}{*}{E} & ALADIN & 0.758(4) & \textbf{0.761(2)} & \textbf{0.761(2)} & \textbf{0.761(2)} \\
 & BLMNII & 0.77(2) & 0.769(3.5) & 0.769(3.5) & \textbf{0.771(1)} \\
 & MLkNNSC & 0.748(4) & \textbf{0.757(1)} & 0.752(2) & 0.751(3) \\
 & NRLMF & 0.786(4) & 0.791(2.5) & 0.791(2.5) & \textbf{0.792(1)} \\
 & WkNN & 0.776(4) & \textbf{0.779(1.5)} & 0.778(3) & \textbf{0.779(1.5)} \\
 & WkNNIR & 0.78(4) & \textbf{0.782(1.5)} & 0.781(3) & \textbf{0.782(1.5)} \\ \midrule
\multirow{6}{*}{DB} & ALADIN & 0.569(4) & \textbf{0.581(1)} & 0.58(2) & 0.579(3) \\
 & BLMNII & 0.433(3) & \textbf{0.448(1)} & 0.444(2) & 0.408(4) \\
 & MLkNNSC & 0.579(4) & \textbf{0.594(1)} & 0.588(2) & 0.586(3) \\
 & NRLMF & 0.585(3) & 0.575(4) & \textbf{0.596(1)} & 0.594(2) \\
 & WkNN & 0.581(2.5) & \textbf{0.594(1)} & 0.581(2.5) & 0.579(4) \\
 & WkNNIR & 0.595(2) & 0.584(4) & \textbf{0.597(1)} & 0.594(3) \\ \bottomrule
\end{tabular}
\end{table}

\begin{table}[]
\centering
\caption{Results of ensemble methods along with their embedded base models in S4 in terms of AUPR. The parenthesis is the rank of each method among all competitors. 
}
\label{ta:en_S4}
\begin{tabular}{@{}cccccc@{}}
\toprule
Dataset & Base Model & Base & ERS & EGS & ELS \\ \midrule
\multirow{6}{*}{NR} & ALADIN & 0.095(3) & \textbf{0.099(1)} & 0.097(2) & 0.094(4) \\
 & BLMNII & 0.135(3.5) & 0.135(3.5) & 0.142(2) & \textbf{0.143(1)} \\
 & MLkNNSC & 0.117(4) & 0.121(3) & \textbf{0.131(1)} & 0.123(2) \\
 & NRLMF & 0.142(4) & 0.149(3) & 0.165(2) & \textbf{0.166(1)} \\
 & WkNN & 0.159(4) & 0.161(3) & \textbf{0.169(1.5)} & \textbf{0.169(1.5)} \\
 & WkNNIR & 0.165(3) & 0.157(4) & \textbf{0.171(1.5)} & \textbf{0.171(1.5)} \\ \midrule
 \multirow{6}{*}{GPCR} & ALADIN & 0.114(4) & 0.116(3) & \textbf{0.121(1)} & 0.12(2) \\
 & BLMNII & 0.121(3.5) & \textbf{0.122(1.5)} & 0.121(3.5) & \textbf{0.122(1.5)} \\
 & MLkNNSC & 0.1(4) & \textbf{0.116(1)} & 0.108(3) & 0.109(2) \\
 & NRLMF & 0.134(3) & 0.132(4) & 0.135(2) & \textbf{0.136(1)} \\
 & WkNN & 0.149(3.5) & 0.149(3.5) & \textbf{0.151(1.5)} & \textbf{0.151(1.5)} \\
 & WkNNIR & 0.158(2.5) & 0.154(4) & 0.158(2.5) & \textbf{0.159(1)} \\ \midrule
\multirow{6}{*}{IC} & ALADIN & 0.206(4) & 0.208(3) & 0.21(2) & \textbf{0.211(1)} \\
 & BLMNII & 0.176(3.5) & 0.176(3.5) & 0.177(2) & \textbf{0.178(1)} \\
 & MLkNNSC & 0.148(4) & \textbf{0.175(1)} & 0.166(2) & 0.165(3) \\
 & NRLMF & 0.215(4) & \textbf{0.223(1)} & 0.217(3) & 0.219(2) \\
 & WkNN & 0.216(3.5) & \textbf{0.22(1)} & 0.216(3.5) & 0.218(2) \\
 & WkNNIR & 0.226(4) & \textbf{0.229(1.5)} & 0.227(3) & \textbf{0.229(1.5)} \\ \midrule
\multirow{6}{*}{E} & ALADIN & 0.128(4) & 0.133(3) & \textbf{0.136(1)} & 0.135(2) \\
 & BLMNII & 0.147(4) & \textbf{0.148(2)} & \textbf{0.148(2)} & \textbf{0.148(2)} \\
 & MLkNNSC & 0.146(4) & \textbf{0.158(1)} & 0.15(3) & 0.155(2) \\
 & NRLMF & 0.198(4) & 0.209(2.5) & 0.209(2.5) & \textbf{0.21(1)} \\
 & WkNN & 0.208(4) & \textbf{0.211(1.5)} & 0.21(3) & \textbf{0.211(1.5)} \\
 & WkNNIR & 0.202(4) & \textbf{0.206(1.5)} & 0.204(3) & \textbf{0.206(1.5)} \\ \midrule
\multirow{6}{*}{DB} & ALADIN & 0.248(4) & \textbf{0.263(1)} & 0.262(2) & 0.261(3) \\
 & BLMNII & 0.064(4) & \textbf{0.075(1)} & 0.073(2.5) & 0.073(2.5) \\
 & MLkNNSC & 0.205(4) & \textbf{0.229(1)} & 0.224(2) & 0.223(3) \\
 & NRLMF & 0.226(3) & 0.203(4) & \textbf{0.23(1.5)} & \textbf{0.23(1.5)} \\
 & WkNN & 0.247(3) & 0.226(4) & 0.248(2) & \textbf{0.25(1)} \\
 & WkNNIR & 0.251(3) & 0.249(4) & 0.252(2) & \textbf{0.253(1)} \\ \bottomrule
\end{tabular}
\end{table}

\end{document}